\documentclass{article}
\usepackage{natbib}

\usepackage[main, final]{neurips_2026}

\usepackage[utf8]{inputenc} 
\usepackage[T1]{fontenc}    
\usepackage{nicefrac}       
\usepackage{microtype}      
\usepackage{xcolor}         
\usepackage{hyperref}
\usepackage{url}
\usepackage{booktabs}       
\usepackage{amsfonts}       
\usepackage{amsmath}
\usepackage{multirow}    
\usepackage{algorithmic} 
\usepackage{pifont}      
\usepackage{tablefootnote} 
\usepackage{enumitem}
\usepackage{graphicx}
\usepackage{amssymb}  
\usepackage{subfigure}
\usepackage{algorithm}
\usepackage{booktabs}
\usepackage{multirow}
\usepackage{graphicx}
\usepackage{xcolor}        
\usepackage{colortbl}      
\usepackage{caption}
\usepackage{booktabs}
\usepackage{tabularx}
\usepackage{fontawesome5}   
\usepackage{twemojis}       

\title{Towards Scalable Context-Aware Single-Cell Spatial Transcriptomics Prediction from Histology Images} 

\author{%
  Zijun Gao$^{1,2}$ \quad Chunbin Gu$^{1}$ \quad
  Jinxi Xiang$^{2}$ \quad Xiangde Luo$^{2}$\textsuperscript{\,\faEnvelope[regular]} \quad Pheng-Ann Heng$^{1}$ \\[0.5em]
  $^{1}$The Chinese University of Hong Kong \quad
  $^{2}$Stanford University School of Medicine
}

\begin{document}

\maketitle
{\renewcommand{\thefootnote}{\faEnvelope[regular]}%
 \footnotetext{Corresponding author: \texttt{luoxd96@stanford.edu}}}
\vspace{-2.5em}
\begin{center}
  \begin{tabular}{@{}l@{\hspace{0.2em}}c@{\hspace{0.2em}}l@{}}
    Code: & \faGithub       & \href{https://github.com/zjgao02/CELLO}{\texttt{github.com/zjgao02/CELLO}} \\[0.2em]
    Model: & \twemoji{1f917} & \href{https://huggingface.co/gaozijun/CELLO}{\texttt{hf.co/gaozijun/CELLO}} \\
    [0.2em]
    Data:  & \twemoji{1f917} & \href{https://huggingface.co/datasets/gaozijun/cello_data}{\texttt{hf.co/datasets/gaozijun/cello\_data}} 
  \end{tabular}
\end{center}

\begin{abstract}
Predicting gene expression from H\&E-stained histology images offers a scalable alternative to costly spatial transcriptomics, yet most existing methods operate at the spot level, where signals from multiple cells are aggregated and critical cellular heterogeneity is obscured. Extending this paradigm to single-cell resolution is non-trivial. Naively applying pathology foundation models faces a scale mismatch: their patch-level representations mix multiple cells, whereas per-cell cropping or resizing distorts morphology and removes local context. Conversely, segmentation-based models without strong pretrained visual encoders often lack the morphological representation capacity needed for accurate molecular prediction and inherit errors from imperfect cell boundary masks. Here, we present CELLO, an efficient end-to-end framework that performs a single pathology foundation model forward pass per image and uses grid sampling to extract location-specific features for all cells simultaneously. We further introduce a distance-decay cross-attention module that refines each cell representation using spatially biased local morphological context. Using 52 public Xenium–H\&E pairs from HEST-1k that span 12 organs and approximately 10 million cells, CELLO improves the average predictive accuracy over the evaluated baselines while reducing the mean whole-slide inference time compared to DeepSpot2Cell, a 14.0× speed-up on average that excludes upstream cell segmentation. Our work establishes a scalable foundation for single-cell gene expression prediction from H\&E images. Code is available at https://github.com/zjgao02/CELLO.

\end{abstract}

\section{Introduction}
\label{introduction}
Hematoxylin and Eosin (H\&E)-stained histology imaging has long been a cornerstone of clinical diagnosis and biomedical research, providing rich morphological details of tissue composition \cite{song2023artificial}. However, it lacks the molecular depth required to fully understand complex disease mechanisms. Spatial transcriptomics (ST) bridges this gap by mapping gene expression profiles to specific spatial locations, revealing the biological heterogeneity of the tumor microenvironment \cite{fu2025spatial, lewis2021spatial}. Despite its immense value, ST remains far less accessible than routine H\&E due to substantial costs and complex experimental protocols \cite{staahl2016visualization}. To democratize spatial molecular profiling, numerous computational methods have been developed to directly infer ST profiles from ubiquitous H\&E images \cite{he2020integrating,xie2023spatially,zeng2022spatial,jia2024thitogene,nonchev2025deepspot,yang2023exemplar,min2024multimodal,zhu2025diffusion}. These pioneering works primarily focus on spot-level predictions, matching the resolution of widely used platforms like 10x Visium, where each micrometer-sized spot captures a mixture of cells \cite{moses2022museum}.

While spot-level prediction provides valuable regional insights, it fundamentally limits our understanding of cellular heterogeneity. A single Visium spot typically encompasses dozens of cells, resulting in an averaged gene expression profile that masks distinct cell types, rare cellular states, and intricate cell-cell interactions. As the field advances, emerging imaging-based platforms like 10x Xenium have pushed spatial resolution to the single-cell level. To truly unlock the potential of digital pathology, there is a pressing need to shift the paradigm of H\&E-to-ST prediction from spot level directly to single-cell level. Achieving this would enable the extraction of highly granular, cell-specific molecular landscapes from standard histology slides without the need for expensive specialized assays.

Recent research has attempted to push beyond spot-level resolution~\cite{bergenstraahle2022super,huang2023single}. Super-resolution approaches, such as iStar~\cite{zhang2024inferring} and scstGCN~\cite{xue2025inferring}, generate spatially refined expression maps at the superpixel level, but they still require ST data as input and do not yield true single-cell profiles. More recent cell-level methods differ mainly in how they construct cell representations. DeepSpot2Cell~\cite{nonchev2025deepspot2cell} leverages the strong visual representation of pathology foundation models (PFMs) by cropping each segmented cell and encoding it independently, then models spots as bags of cells under weak spot-level supervision. However, this strategy is a naive adaptation of patch-level PFMs to irregular cell-level targets: per-cell cropping and resizing can distort cellular morphology, remove local microenvironmental context, and fail to align naturally with the patch-level representations learned during pretraining. It also requires an independent PFM forward pass for every cell, which is computationally prohibitive for WSIs containing hundreds of thousands to millions of cells. Another line of work, exemplified by GHIST~\cite{fu2025spatial}, constructs more cell-aware representations from segmentation masks, which can better preserve cell-level spatial structure. However, such models do not fully benefit from the strong morphology representations learned by modern PFMs, and their predictions remain tightly coupled to the quality of upstream cell segmentation, as illustrated in Fig.\ref{fig:main}.

These limitations highlight the central challenge of single-cell H\&E-to-ST prediction: obtaining accurate, scalable, and context-aware cell embeddings. To address this, we propose \textbf{CELLO} (\underline{\textbf{CEL}}l-\underline{\textbf{L}}evel gene expression prediction from hist\underline{\textbf{O}}logy), an efficient end-to-end framework that predicts gene expression at cell locations from H\&E images. CELLO performs a single pathology foundation model forward pass per image and uses grid sampling to extract location-specific features for all cells simultaneously, avoiding per-cell cropping, boundary-mask dependence, and the ill-posed disentanglement of cell profiles from aggregated spot-level supervision. A distance-decay cross-attention module further refines each cell representation with spatially biased local morphological context, as illustrated in Fig.\ref{fig:main}D.

In summary, our main contributions are as follows:
\begin{itemize}[noitemsep, topsep=0pt, leftmargin=1.2em]

    \item We present a scalable end-to-end framework for single-cell H\&E-to-ST prediction that directly trains on paired single-cell ST and H\&E images, rather than relying on weakly supervised spot-level expression.

    \item We drastically reduce computational cost and mitigate the cascading effects of segmentation inaccuracies using grid sampling, which preserves a strong visual representation of PFM.

    \item We explore multiple context fusion strategies and demonstrate that distance-decay cross attention most effectively incorporates spatially biased local morphological context for strong cell representation.

    \item Extensive experiments on 52 paired Xenium--WSI samples across 12 organs from HEST-1k show that CELLO achieves state-of-the-art predictive performance and substantially improved computational efficiency that excludes upstream cell segmentation.
\end{itemize}

\begin{figure}[htbp]
  \centering
  \includegraphics[width=\linewidth]{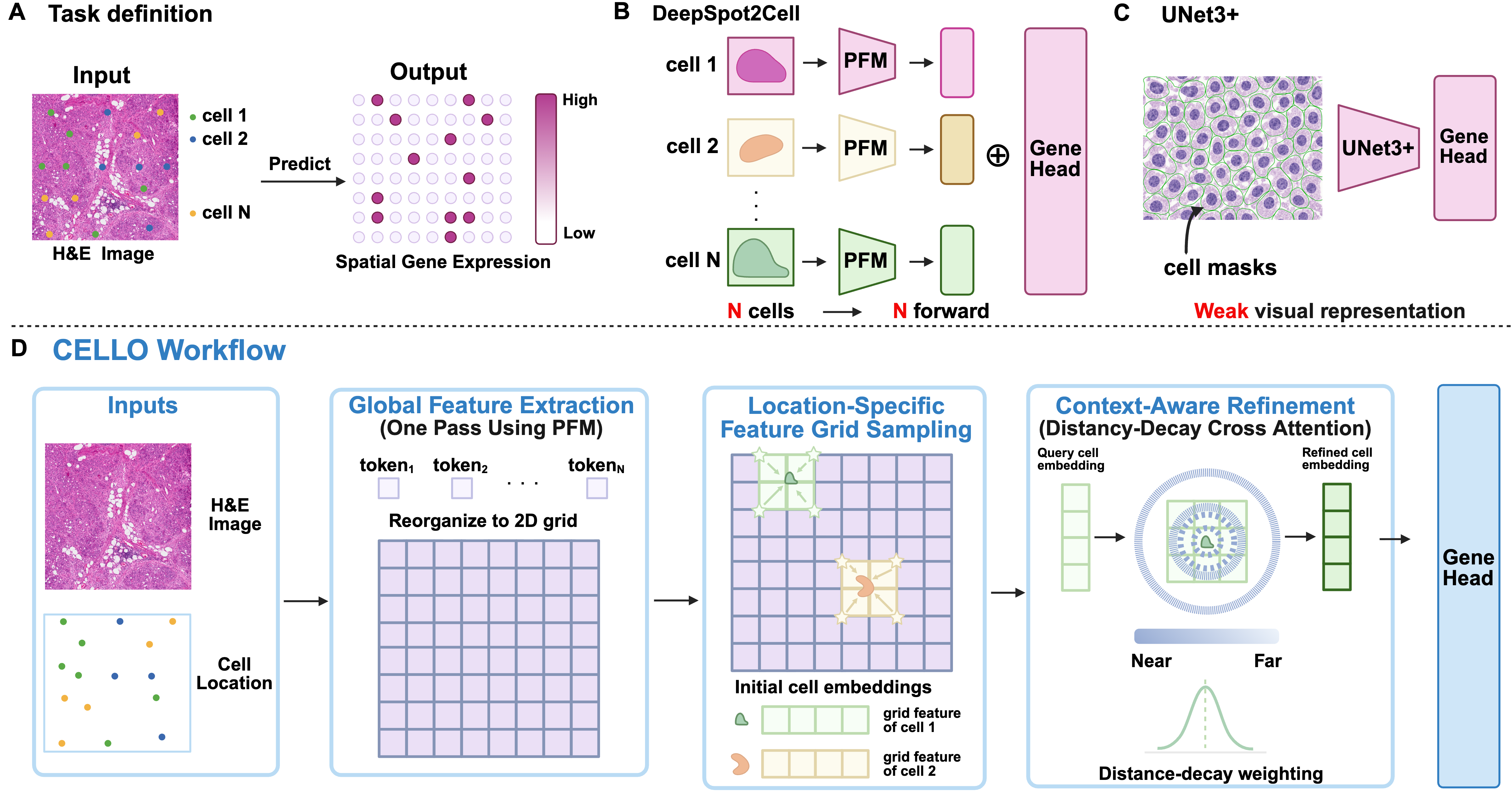}
  \caption{\textbf{(A)} Task definition: predicting single-cell spatial gene expression from H\&E images with cell locations. \textbf{(B)} PFM-based naive cropping methods such as DeepSpot2Cell require $N$ per-cell forward passes and distort local morphology. \textbf{(C)} Cell mask based UNet3+ lacks strong pretrained visual representations. \textbf{(D)} CELLO performs a single PFM forward pass and uses grid sampling to extract initial feature for all cells simultaneously. A context-aware cross-attention module with 2D RoPE and Gaussian distance-decay bias captures local cellular context and cell--cell interactions.}

  \label{fig:main}
\end{figure}

\section{Related Work}

\subsection{Spatial Transcriptomics Prediction From H\&E Images
}
Early regression-based methods extract patch-level visual features and map
them to expression vectors via transfer
learning~\cite{he2020integrating,pang2021leveraging,zeng2022spatial}, with later improvements
incorporating multi-resolution context~\cite{chung2024accurate,wang2024m2ort} or
retrieval-based strategies~\cite{xie2023spatially,yang2023exemplar}.
Beyond regression, recent work has explored contrastive
alignment~\cite{xie2023spatially}, denoising diffusion~\cite{zhu2025diffusion}, and flow
matching~\cite{huang2025scalable} to better capture the stochastic and
multimodal nature of gene expression.
Despite these methodological advances, nearly all existing approaches operate
at the \textbf{spot level}, predicting a single expression vector that aggregates signal from hundreds of cells.
Several recent efforts attempt to push beyond
spot-level granularity, broadly following two strategies:
\textbf{(1) Super-resolution-based deconvolution~\cite{xue2025inferring,zhang2024inferring,zhao2020bayesspace}.} They use hierarchical ViT or graph convolutional networks to
produce high-resolution expression maps. However, their outputs are high-dimensional superpixel expression maps rather than precise cell transcriptomic profiles, and they still require custom post-processing to approximate.
\textbf{(2) Spot-level weakly supervised approaches} such as
DeepSpot2Cell~\cite{nonchev2025deepspot2cell} distributes
spot-level expression to individual cells detected within each spot.
These methods cause prohibitive inference costs that prevent
scaling to whole-slide images containing millions of cells. GHIST~\cite{fu2025spatial} builds single cell-aware representations from segmentation masks, better preserving cell-level spatial structure. However, it does not fully exploit the rich morphological features learned by modern PFMs.
\textbf{CELLO} addresses these limitations by directly training on true pairs of single-cell ST and H\&E images and employing an end-to-end, scalable architecture, enabling efficient scaling to whole-slide-level (WSI) analysis.

\subsection{Computational Pathology Foundation Models}
Foundation models in computational pathology have made great progress in generating patch-level embeddings that transfer across diverse downstream tasks~\cite{jaume2024hest}.
They are trained on millions of histology images via
self-supervised learning objectives including DINO~\cite{zhang2022dino},
image-text contrastive learning~\cite{radford2021learning}, and masked image
modeling and have demonstrated that rich morphological and tissue-level information can be captured in compact feature representations ~\cite{chen2025visual}. Notable examples include UNI~\cite{chen2024towards}, Virchow2~\cite{zimmermann2024virchow2}, 
H-Optimus-0~\cite{saillard2024h}. The latter two are scaled to larger ViT architectures and trained on broader data.
In this work, we treat the foundation models as a feature extractor and systematically evaluate them to assess how encoder choice affects cell-level expression prediction.

\section{Method}

\begin{figure}[htbp]
  \centering
  \includegraphics[width=\linewidth]{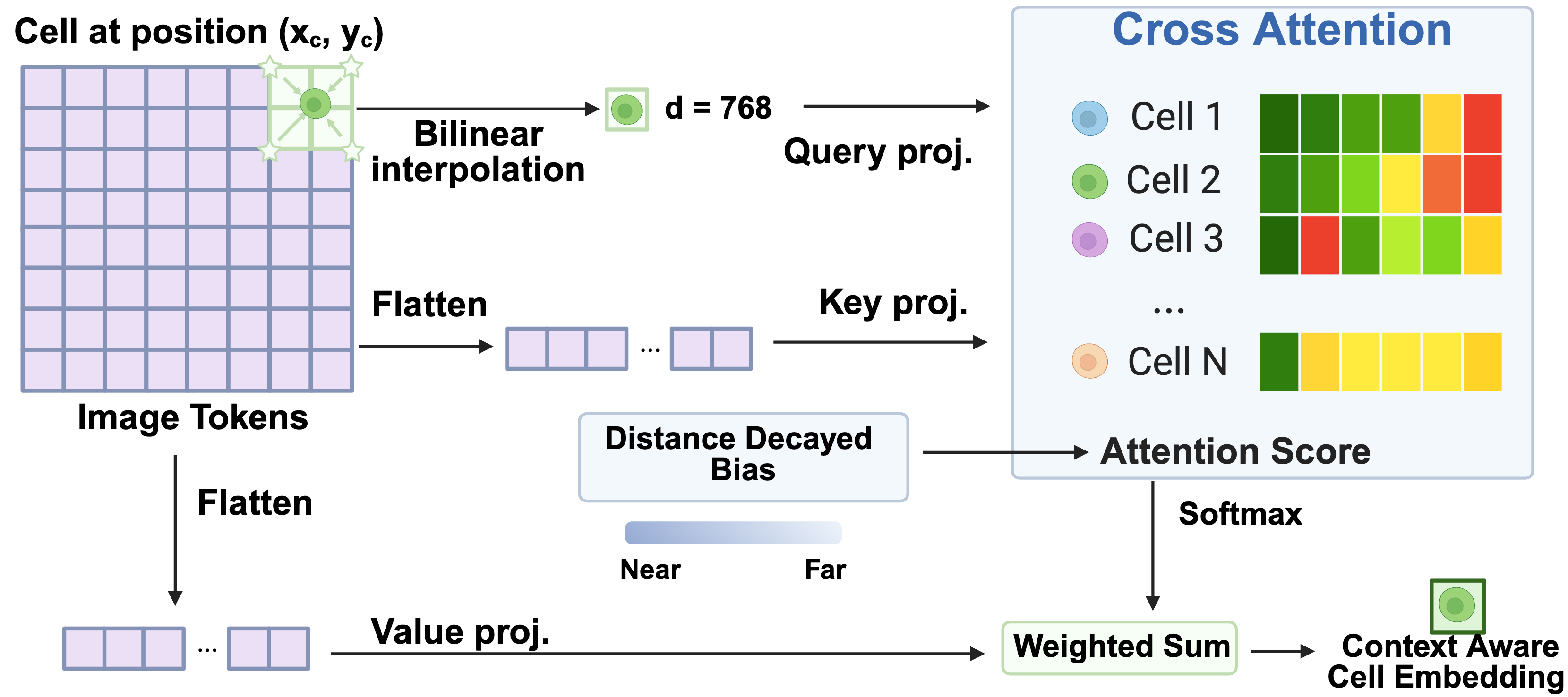}
  \caption{
Given the 2D visual token map from a single PFM forward pass, CELLO first queries a location-specific cell feature by bilinear interpolation. The cell query is then refined by distance-decay cross-attention over patch tokens, where spatially nearby tokens receive a larger prior.
This produces a context-aware cell embedding that preserves precise cell-location information while incorporating local morphological context.}

  \label{fig:main}
\end{figure}

\subsection{Problem Formulation}

Let $\mathbf{V} \in \mathbb{R}^{L \times L \times 3}$ denote an image extracted from a H\&E-stained whole-slide image, where $L$ is the patch size in pixels. Within each patch, we observe a variable number of cells $\{c_i\}_{i=1}^{N}$, each associated with a spatial coordinate $\mathbf{s}_i = (x_i, y_i) \in \mathbb{R}^2$ obtained via spatial transcriptomics or cell segmentation, and a ground-truth gene expression vector $\mathbf{x}_i \in \mathbb{R}_{\geq 0}^{G}$ over a union gene vocabulary of size $G$.

Prior work typically predicts one expression vector per spot, i.e., $f:\mathbf{V}\mapsto\mathbf{x}$, which cannot resolve cell-level heterogeneity within the patch. We instead formulate H\&E-to-ST prediction at single-cell resolution. Given an image tile $\mathbf{V}$ and cell locations $\mathcal{S}=\{\mathbf{s}_i\}_{i=1}^{N}$, we learn
\begin{equation}
    f_\theta: (\mathbf{V},\, \mathbf{s}_i) \;\longmapsto\; \hat{\mathbf{x}}_i \in \mathbb{R}_{\geq 0}^{G}, \quad i = 1, \ldots, N,
\end{equation}
where $\hat{\mathbf{x}}_i$ denotes the predicted gene expression vector for cell $i$. This formulation requires extracting spatially precise visual features at each cell location while preserving local morphological context, which motivates our position-aware querying and context-fusion design.

\subsection{Visual Feature Extraction}


We use a pretrained pathology foundation model $f_{\text{enc}}$ as the visual backbone. Given an H\&E image patch $\mathbf{V}$, the encoder produces a CLS token $\mathbf{z}_{\texttt{cls}}\in\mathbb{R}^{d}$ and $M$ spatial tokens $[\mathbf{t}_1,\ldots,\mathbf{t}_M]\in\mathbb{R}^{M\times d}$ from non-overlapping $p\times p$ sub-patches. Because these tokens correspond to a regular image grid, we reshape them into a 2D feature map $\mathbf{T}\in\mathbb{R}^{L_g\times L_g\times d}$, where $L_g=L/p$ and $M=L_g^2$. This preserves the geometric correspondence between visual tokens and image regions, enabling continuous cell-location querying.

\subsection{Position-Aware Cell Query Construction}

A central design goal is to construct a cell-level representation that is anchored to the cell's precise spatial location within the patch, rather than being tied to a discrete token grid. We achieve this through a two-stage process.

\paragraph{Bilinear Query Initialization.}
Each cell's spatial coordinate $\mathbf{s}_i$ is recorded in the global whole-slide coordinate system. We first convert it to a patch-local coordinate by subtracting the top-left corner $\mathbf{o} \in \mathbb{R}^2$ of the patch: $\mathbf{s}_i^{\text{loc}} = \mathbf{s}_i - \mathbf{o}$, yielding a position in $[0, L)^2$. We then normalize to the $[-1,1]$ range required by bilinear sampling:
\begin{equation}
    \tilde{\mathbf{s}}_i = \frac{2\,\mathbf{s}_i^{\text{loc}}}{L} - 1.
\end{equation}
The initial cell query is extracted from the spatial feature map $\mathbf{T}$ via differentiable bilinear interpolation~\cite{jaderberg2015spatial}:
\begin{equation}
    \mathbf{q}_i^{(0)} = \texttt{GridSample}\!\bigl(\mathbf{T},\;\tilde{\mathbf{s}}_i\bigr) \in \mathbb{R}^d,
\end{equation}
which produces a continuous, sub-patch feature by smoothly interpolating information from the four nearest spatial tokens surrounding the cell.

\paragraph{Cross-Attention with 2D RoPE and Distance Decay.}
The initial query $\mathbf{q}_i^{(0)}$ is then refined by attending over all spatial tokens via multi-head cross-attention. To inject explicit spatial structure, we augment the attention mechanism with two complementary positional inductive biases.

First, we apply 2D RoPE~\cite{su2024roformer} to queries and keys. It encodes relative spatial relationships between the cell query and each image token directly into the attention logits without additional parameters. For cell queries, we encode positions in the same coordinate system as visual tokens. Specifically, cell pixel coordinates are divided by the visual token stride to obtain continuous token-grid coordinates, while visual keys use token-center grid coordinates. This avoids feeding raw pixel coordinates into RoPE and aligns the positional scale between cell queries and image tokens.

Second, we add a \emph{distance-decay bias} to the pre-softmax attention scores:
\begin{equation}
    b_{i,m} = -\frac{\|\mathbf{s}_i^{\text{loc}} - \mathbf{c}_m\|^2}{2p^2},
\end{equation}
where $\mathbf{c}_m \in \mathbb{R}^2$ is the pixel-space center of the $m$-th spatial token within the patch, computed as $\mathbf{c}_m = (p_m^x \cdot p + p/2,\; p_m^y \cdot p + p/2)$ with $(p_m^x, p_m^y)$ being the grid index of token $m$. Both $\mathbf{s}_i^{\text{loc}}$ and $\mathbf{c}_m$ are in the same patch-local pixel coordinate system, and the normalization by $p^2$ makes the decay scale relative to the token receptive field size. Finally, the output is layer-normalized to yield the cell embedding $\mathbf{z}_i \in \mathbb{R}^d$

\subsection{Gene Expression Prediction and Training Objectives}

The final cell representation $\mathbf{z}_i$ is decoded into a predicted gene expression vector through a two-layer MLP with ReLU activation: $\hat{\mathbf{x}}_i = h_{\text{cell}}(\mathbf{z}_i') \in \mathbb{R}_{\geq 0}^{G}$. The model is trained with a weighted combination of cell-level and patch-level Huber losses~\cite{huber1992robust}:
\begin{equation}
    \mathcal{L} = \mathcal{L}_{\text{cell}} + \lambda\, \mathcal{L}_{\text{patch}},
\end{equation}
where $\mathcal{L}_{\text{cell}}$ averages per-gene Huber loss over all cells and active genes. The auxiliary patch-level loss $\mathcal{L}_{\text{patch}}$ is obtained by predicting the aggregate expression within the patch, with the patch-level target defined as the sum of all cell-level expression vectors in that patch, i.e., $\mathbf{x}_{\text{patch}}=\sum_{i=1}^{N}\mathbf{x}_i$. We use Huber loss because CELLO is formulated as supervised regression on normalized gene expression values and is evaluated by Pearson correlation, rather than as generative modeling of raw transcript counts. Compared with mean squared error, Huber loss is less sensitive to heavy-tailed expression values and occasional outliers.

\section{Experiments}
In this section, we comprehensively evaluate our proposed method. Specifically, we present: (1) benchmark results on test datasets across 12 different organs from human; (2) ablation studies analyzing the impact of different pretrained encoders and context integration; (3) a computational efficiency comparison demonstrating the superior scalability of our approach over baseline methods; and (4) visualizations through representative case studies.

\paragraph{Evaluation Metrics}
We train a unified gene prediction head over a union vocabulary of 1,915 genes; each sample's panel contains 280–480 genes (Appendix~\ref{app:dataset}). Since individual Xenium panels may contain different measured genes, the training loss and evaluation metrics are computed only over the genes observed in the corresponding sample. Our evaluation metric is the mean Pearson Correlation Coefficient (PCC). Following widely used gene selection protocols, we selected three gene sets: top-k most predictive genes (denoted as MPG)~\cite{nonchev2025deepspot}, top-k most predictive genes from the highly variant genes ordered in mean
(denoted as HVG) and spatially variable genes (denoted as SVG)~\cite{he2020integrating}. In the main text, we will only show HVG; the other two gene subsets are shown in the  Appendix~\ref{app:full_gene_results}. The MPG, HVG, and SVG top-$k$ sets are used only as evaluation subsets of the predicted gene vocabulary and do not affect model training. We also report performance over the full gene set in Appendix~\ref{app:full_gene_results} to provide a more comprehensive assessment beyond top-$k$ evaluation subsets.

\subsection{10X-Xenium-52 Benchmark}

\paragraph{Dataset and Preprocessing}
We evaluate CELLO and baseline methods on a dataset comprising 52 paired Whole Slide Images (WSIs) and Spatial Transcriptomics (ST) samples sequenced via the Xenium platform from HEST-1k~\cite{jaume2024hest}. The dataset spans 12 distinct organs across three health conditions (cancer, diseased, and healthy), with significant variations ranging from 16,429 to 882,692 cells per sample. During preprocessing, WSIs are cropped into 224×224 patches, yielding between 3,465 and 93,515 patches per sample. The total number of cells is approximately 10M. We log-transformed the gene expression following~\cite{zhu2025diffusion}.

To rigorously assess the model's generalization capability without data leakage, we form a 4-sample validation set by sampling from the most abundant organs (lung, breast, bowel). The test set is carefully designed to evaluate model performance under two distinct settings:

\begin{itemize}[noitemsep, topsep=0pt, leftmargin=1.2em]

\item \textbf{In-Distribution (ID) Setting:} Evaluates performance on six organ types (Lung, Breast, Bowel, Skin, Pancreas, and Lymphoid). For these samples, both the organ types and their associated health conditions were observed during training, assessing the model's capacity to exploit statistical regularities within well-represented categories.
\item \textbf{Out-of-Distribution (OOD) Setting:} provides case studies of transfer to organs or conditions absent from training. This includes four \textbf{completely unseen} organs (Brain, Bone, Heart, and Ovary), each represented by a single WSI reserved exclusively for testing. The other two organs (Kidney and Liver) were seen during training, but their specific health conditions present at test time were never encountered.
\end{itemize}

All remaining WSIs are allocated to the training set. Dataset details are provided in Appendix~\ref{app:dataset}

\paragraph{Baseline Methods}
Single-cell H\&E-to-ST prediction remains under-explored, and only a few recent methods explicitly study how to construct cell embeddings from histology. The first strategy uses segmentation-aware CNN features, as in GHIST~\cite{fu2025spatial}, where UNet3+ is used to extract cell representations from cell masks. This design preserves mask-defined cell structure but lacks strong visual representation without leveraging modern PFMs. The second strategy uses PFM features from per-cell crops, as in DeepSpot2Cell~\cite{nonchev2025deepspot2cell}. While this benefits from strong pretrained visual representations, it requires cell cropping and separate feature extraction for each cell, which is computationally expensive and can distort cell morphology or remove local context. Finally, to isolate the contribution of contextual fusion, we include a grid-sampling baseline, denoted as \emph{Grid}, which queries cell features from the shared patch-level feature map using cell locations but does not apply distance-decay cross-attention. Training details are provided in Appendix~\ref{app:training_details}.

\paragraph{Results}
\begin{table}[htbp]
    \centering
    
    \caption{\textbf{HVG results on the 10X-Xenium-52 benchmark.} ID tissues are shown on the left and OOD tissues on the right. Higher PCC values are better. The best results within each tissue are highlighted in \textbf{bold}.}
    \label{tab:main_hvg}
    \scriptsize
    \setlength{\tabcolsep}{2.6pt}
    \resizebox{\textwidth}{!}{%
    \begin{tabular}{llccc@{\hspace{4pt}}!{\vrule width 0.35pt}@{\hspace{4pt}}llccc}
    \toprule
    \multicolumn{5}{c}{\textbf{In-distribution (ID)}} &
    \multicolumn{5}{c}{\textbf{Out-of-distribution (OOD)}} \\
    \cmidrule(lr){1-5} \cmidrule(lr){6-10}
    Tissue & Model & HVG-10 & HVG-50 & HVG-200 &
    Tissue & Model & HVG-10 & HVG-50 & HVG-200 \\
    \midrule

    \multirow{4}{*}{Lung} & UNet3+        & 0.4293 & 0.2177 & 0.0824
    & \multirow{4}{*}{Kidney} & UNet3+        & 0.1828 & 0.1189 & 0.0367 \\
    & DeepSpot2Cell & 0.4082 & 0.2264 & 0.0763
    & & DeepSpot2Cell & 0.3303 & 0.1803 & 0.0610 \\
    & Grid          & 0.5481 & 0.3730 & 0.1455
    & & Grid          & 0.4728 & \textbf{0.3147} & \textbf{0.1235} \\
    & \textbf{CELLO} & \textbf{0.5537} & \textbf{0.3827} & \textbf{0.1492}
    & & \textbf{CELLO} & \textbf{0.5515} & 0.2743 & 0.1078 \\
    \cmidrule(lr){1-5} \cmidrule(lr){6-10}

    \multirow{4}{*}{Breast} & UNet3+        & 0.3078 & 0.1910 & 0.0546
    & \multirow{4}{*}{Heart} & UNet3+        & 0.2141 & 0.1209 & 0.0430 \\
    & DeepSpot2Cell & 0.5228 & 0.3583 & 0.1340
    & & DeepSpot2Cell & 0.2018 & 0.1106 & 0.0349 \\
    & Grid          & 0.6325 & 0.4932 & 0.2318
    & & Grid          & 0.3490 & \textbf{0.1868} & \textbf{0.0672} \\
    & \textbf{CELLO} & \textbf{0.6736} & \textbf{0.5243} & \textbf{0.2541}
    & & \textbf{CELLO} & \textbf{0.4070} & 0.1469 & 0.0258 \\
    \cmidrule(lr){1-5} \cmidrule(lr){6-10}

    \multirow{4}{*}{Bowel} & UNet3+        & 0.3868 & 0.2615 & 0.1058
    & \multirow{4}{*}{Liver} & UNet3+        & 0.3957 & 0.2616 & 0.0839 \\
    & DeepSpot2Cell & 0.6358 & 0.4163 & 0.1494
    & & DeepSpot2Cell & 0.4622 & 0.3410 & 0.1218 \\
    & Grid          & 0.6742 & 0.4821 & 0.2125
    & & Grid          & 0.5217 & 0.3932 & 0.1570 \\
    & \textbf{CELLO} & \textbf{0.7732} & \textbf{0.6599} & \textbf{0.3313}
    & & \textbf{CELLO} & \textbf{0.6609} & \textbf{0.5165} & \textbf{0.1707} \\
    \cmidrule(lr){1-5} \cmidrule(lr){6-10}

    \multirow{4}{*}{Skin} & UNet3+        & 0.3806 & 0.1985 & 0.0671
    & \multirow{4}{*}{Bone} & UNet3+        & 0.1042 & 0.0553 & 0.0122 \\
    & DeepSpot2Cell & 0.5219 & 0.2689 & 0.0854
    & & DeepSpot2Cell & 0.1351 & 0.0638 & 0.0164 \\
    & Grid          & 0.6444 & 0.3520 & 0.1232
    & & Grid          & 0.1626 & 0.0837 & 0.0244 \\
    & \textbf{CELLO} & \textbf{0.7672} & \textbf{0.4503} & \textbf{0.1302}
    & & \textbf{CELLO} & \textbf{0.2113} & \textbf{0.1024} & \textbf{0.0290} \\
    \cmidrule(lr){1-5} \cmidrule(lr){6-10}

    \multirow{4}{*}{Pancreas} & UNet3+        & 0.2761 & 0.1709 & 0.0589
    & \multirow{4}{*}{Brain} & UNet3+        & 0.2021 & 0.0875 & 0.0256 \\
    & DeepSpot2Cell & 0.3123 & 0.1977 & 0.0580
    & & DeepSpot2Cell & 0.2054 & 0.1011 & 0.0275 \\
    & Grid          & 0.3382 & 0.2197 & \textbf{0.0864}
    & & Grid          & \textbf{0.4254} & 0.1921 & 0.0596 \\
    & \textbf{CELLO} & \textbf{0.4181} & \textbf{0.2640} & 0.0767
    & & \textbf{CELLO} & 0.4044 & \textbf{0.2355} & \textbf{0.0731} \\
    \cmidrule(lr){1-5} \cmidrule(lr){6-10}

    \multirow{4}{*}{Lymphoid} & UNet3+        & 0.3606 & 0.1870 & 0.0579
    & \multirow{4}{*}{Ovary} & UNet3+        & 0.4089 & 0.1929 & 0.0572 \\
    & DeepSpot2Cell & 0.3451 & 0.1959 & 0.0520
    & & DeepSpot2Cell & 0.3957 & 0.1840 & 0.0526 \\
    & Grid          & 0.5029 & 0.3001 & 0.0851
    & & Grid          & 0.5184 & 0.2773 & 0.0857 \\
    & \textbf{CELLO} & \textbf{0.6667} & \textbf{0.3528} & \textbf{0.0891}
    & & \textbf{CELLO} & \textbf{0.6258} & \textbf{0.4049} & \textbf{0.1435} \\
    \midrule

    \multirow{4}{*}{Average} & UNet3+        & 0.3569 & 0.2044 & 0.0711
    & \multirow{4}{*}{Average} & UNet3+        & 0.2513 & 0.1395 & 0.0431 \\
    & DeepSpot2Cell & 0.4577 & 0.2773 & 0.0925
    & & DeepSpot2Cell & 0.2884 & 0.1635 & 0.0524 \\
    & Grid          & 0.5567 & 0.3700 & 0.1474
    & & Grid          & 0.4083 & 0.2413 & 0.0862 \\
    & \textbf{CELLO} & \textbf{0.6421} & \textbf{0.4390} & \textbf{0.1718}
    & & \textbf{CELLO} & \textbf{0.4768} & \textbf{0.2801} & \textbf{0.0916} \\

    \bottomrule
    \end{tabular}%
    }
\end{table}

As shown in Table \ref{tab:main_hvg}, CELLO substantially outperforms UNet3+ and DeepSpot2Cell in both ID and OOD settings, demonstrating the effectiveness of constructing cell embeddings through location-specific querying over shared PFM features. In the ID setting, CELLO improves over the strongest baseline, DeepSpot2Cell, by 58.5\% on HVG PCC-50. This advantage becomes even larger in the OOD setting, where CELLO achieves a 71.3\% relative improvement on HVG PCC-50, indicating stronger robustness under unseen tissue and disease conditions. In addition, the comparison with \emph{Grid}, an ablated version of CELLO without the distance-decay cross-attention module, further validates the importance of contextual refinement. CELLO improves over Grid by 18.6\% and 16.2\% on HVG PCC-50 in the ID and OOD settings, respectively, showing that spatially biased cross-attention provides complementary local morphological context beyond location-based feature querying alone.

\subsection{Ablation Study}
In this section, we conduct comprehensive ablation studies to validate the key design choices of CELLO. Specifically, we first investigate the impact of selecting different pathology foundation models as our vision encoder. Furthermore, we explore a different context for fusion to identify which context truly matters in this task.

\begin{figure}[t!]
  \centering
  \includegraphics[width=\linewidth]{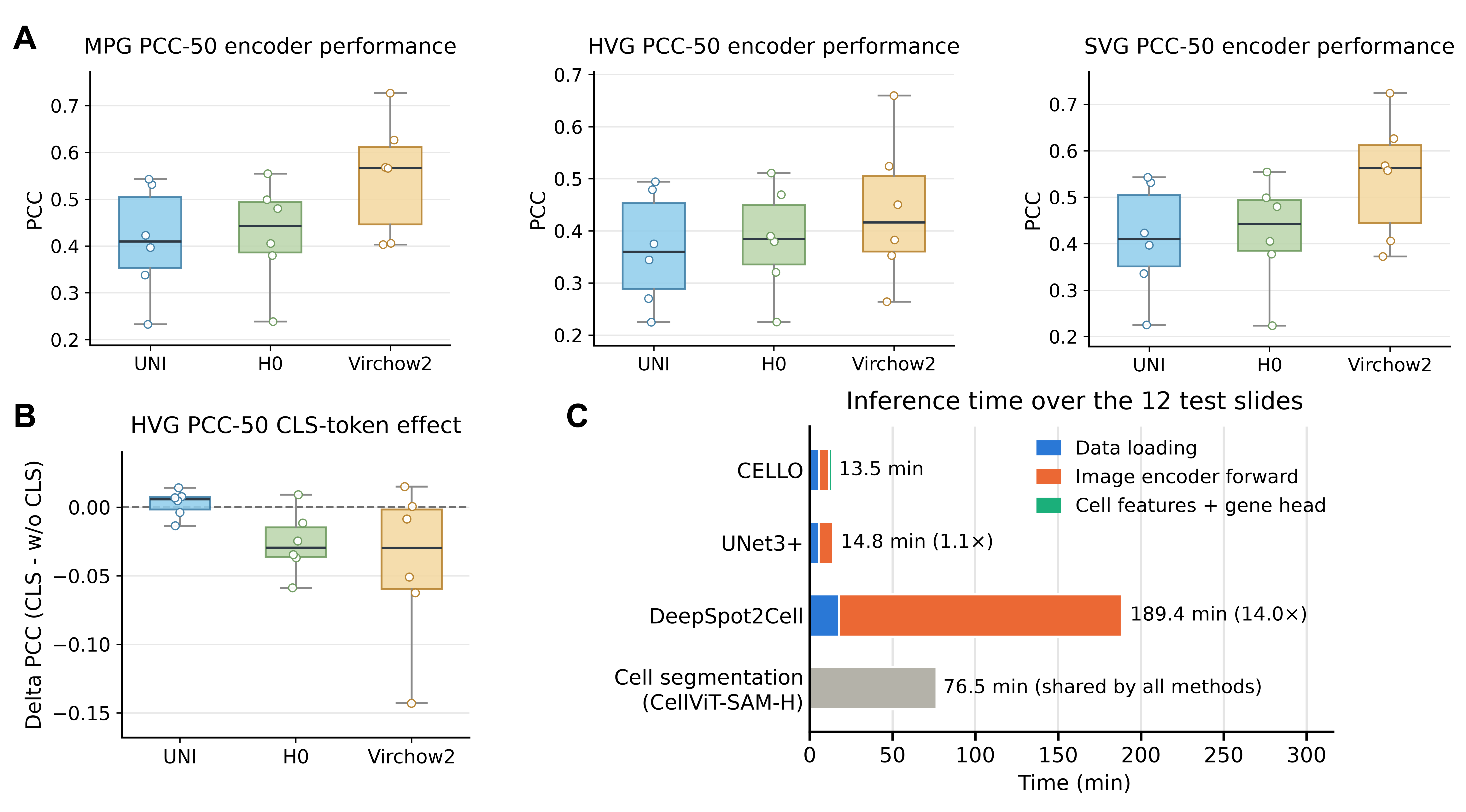}
  \caption{\textbf{Ablation and efficiency analysis of CELLO.}
\textbf{(A)} Performance comparison of different vision encoders on ID settings. Columns show, from left to right: MPG-10, HVG-10, SVG-10.
\textbf{(B)} Effect of CLS-token fusion. Mean $\Delta$PCC is computed as PCC (with cls) - PCC (w/o cls), averaged across different vision encoders.
\textbf{(C)} Total inference time over the 12 test slides, split into data loading, image-encoder forward pass, and cell features with gene head. Cell segmentation (CellViT-SAM-H), required by all methods, is shown separately. CELLO is 14.0$\times$ faster than DeepSpot2Cell.}

  \label{fig:ablation}
\end{figure}

\paragraph{Choice of Foundation Models}
To ensure a fair comparison among different pre-trained pathology foundation models and to prevent the introduction of potential confounding factors, we abstain from using any data augmentation techniques in these experiments. Specifically, we evaluate three representative models: UNI~\cite{chen2024towards}, Virchow2~\cite{zimmermann2024virchow2}, 
H-Optimus-0~\cite{saillard2024h}. These models vary in their architectural specifications: UNI utilizes a patch size of 16×16 with a feature dimension of 1024, whereas both Virchow2 and H-optimus-0 employ a 14×14 patch size, yielding feature dimensions of 1280 and 768, respectively.

As illustrated in Fig.~\ref{fig:ablation}A, across MPG, HVG, and SVG evaluation sets, Virchow2 consistently achieves the best performance among the three evaluated encoders. On PCC-50, Virchow2 obtains average scores of 0.5494, 0.4390, and 0.5425 for MPG, HVG, and SVG, respectively, outperforming H0 by 28.9\%, 14.7\%, and 28.1\%, and UNI by 33.7\%, 20.3\%, and 32.6\%. H0 also provides modest but consistent gains over UNI across all three gene sets. These results suggest that stronger pathology foundation representations substantially improve the quality of cell embeddings, especially for predictive and spatially variable genes, and motivate our use of Virchow2 as the default visual encoder in CELLO.

\paragraph{What Context Really Matters}
A natural question arising from our architecture design is what type of contextual information truly matters for single-cell gene expression prediction. Our baseline model primarily captures context through local cell-cell interactions within the spatial neighborhood of each target patch, reflecting the biological intuition that a cell's transcriptomic state is shaped by its immediate microenvironment. However, it remains unclear whether supplementing this local context with broader, slide-level information could further enhance prediction. To investigate this, we conduct experiments that augment final cell features with the CLS token from the vision encoder, which encodes a global summary of the entire histology image. 
Specifically, for each cell embedding $\mathbf{z}_i$ and the corresponding patch CLS token $\mathbf{z}_{\texttt{cls}}$, we learn a scalar gate to adaptively fuse local cell-level and global patch-level context:
\[
    g_i = \sigma\!\left(\phi_g\!\left([\mathbf{z}_i;\mathbf{z}_{\texttt{cls}}]\right)\right), 
    \qquad
    \tilde{\mathbf{z}}_i = g_i \mathbf{z}_i + (1-g_i)\mathbf{z}_{\texttt{cls}},
\]
where $[\cdot;\cdot]$ denotes concatenation, $\phi_g$ is a two-layer MLP with GELU activation, and $g_i\in(0,1)$ controls the contribution of local versus global context for cell $i$. Then $\tilde{\mathbf{z}}_i$ will be served as final cell embedding instead of being used directly for calculating $\mathcal{L}_{\text{patch}}$.

If high-level contextual information were beneficial, one would expect consistent improvements from this augmentation. Instead, as we show in Fig.~\ref{fig:ablation}B, incorporating the CLS token consistently degrades performance, providing strong empirical evidence that single-cell gene expression prediction is predominantly driven by fine-grained, local morphological context rather than abstract global semantics. We provide detailed ID and OOD results in Appendix~\ref{app:ablation}.

\subsection{Computational Efficiency}
A critical consideration for practical deployment on whole-slide images (WSIs) is inference speed. Existing methods suffer from inherent scalability bottlenecks. DeepSpot2Cell requires a separate forward pass through a frozen pathology foundation model (PFM) for each individual cell tile, in addition to the spot tile and neighboring spot tiles. Since a single WSI can contain hundreds of thousands to millions of cells, this per-cell PFM inference introduces substantial computational overhead that scales linearly with cell count, making it prohibitively expensive for large-scale slides.

Given cell locations, CELLO and UNet3+ take comparable time on the 12 test slides: CELLO needs $67.5\pm48.8$~s per slide and UNet3+ $73.9\pm52.1$~s. Across slides, the UNet3+/CELLO time ratio ranges from 0.99 to 1.21 (median 1.12$\times$), and it is smallest on the largest slides. Most of CELLO's cost is the single Virchow2 forward pass per patch, which takes 81.8\% of its GPU time; grid sampling, cross-attention, CLS fusion and the gene head together add less than 20\%. When cell segmentation with CellViT-SAM-H is included (382.5~s per slide on average), the two pipelines are on par (median 1.02$\times$). Compared with dense mask-based models such as UNet3+, the advantage of CELLO is therefore higher accuracy at a similar cost rather than speed; the large speed-up applies to methods that run the foundation model once per cell, as discussed next. 

We also compare CELLO with DeepSpot2Cell, which runs the foundation model once per cell (Fig.~\ref{fig:ablation}C). On the 12 test slides, DeepSpot2Cell needs $947.2\pm1068.3$~s per slide, 14.0$\times$ the time of CELLO on average (median 11.7$\times$, range 3.5--26.3$\times$). The speed-up grows with cell density (Spearman $\rho=0.97$ with the mean number of cells per patch), because DeepSpot2Cell runs one forward pass per cell whereas CELLO runs one per patch: 114,429 patches versus 2,996,926 cell crops in total. Including cell segmentation, CELLO remains 2.5$\times$ faster (median over slides).

These results demonstrate that CELLO not only exceeds the prediction accuracy of existing approaches but does so with substantially greater computational efficiency, making it a practical solution for large-scale spatial transcriptomics inference from H\&E images.
\subsection{Case Study}
\begin{figure}[htbp]
  \centering
  \includegraphics[width=\linewidth]{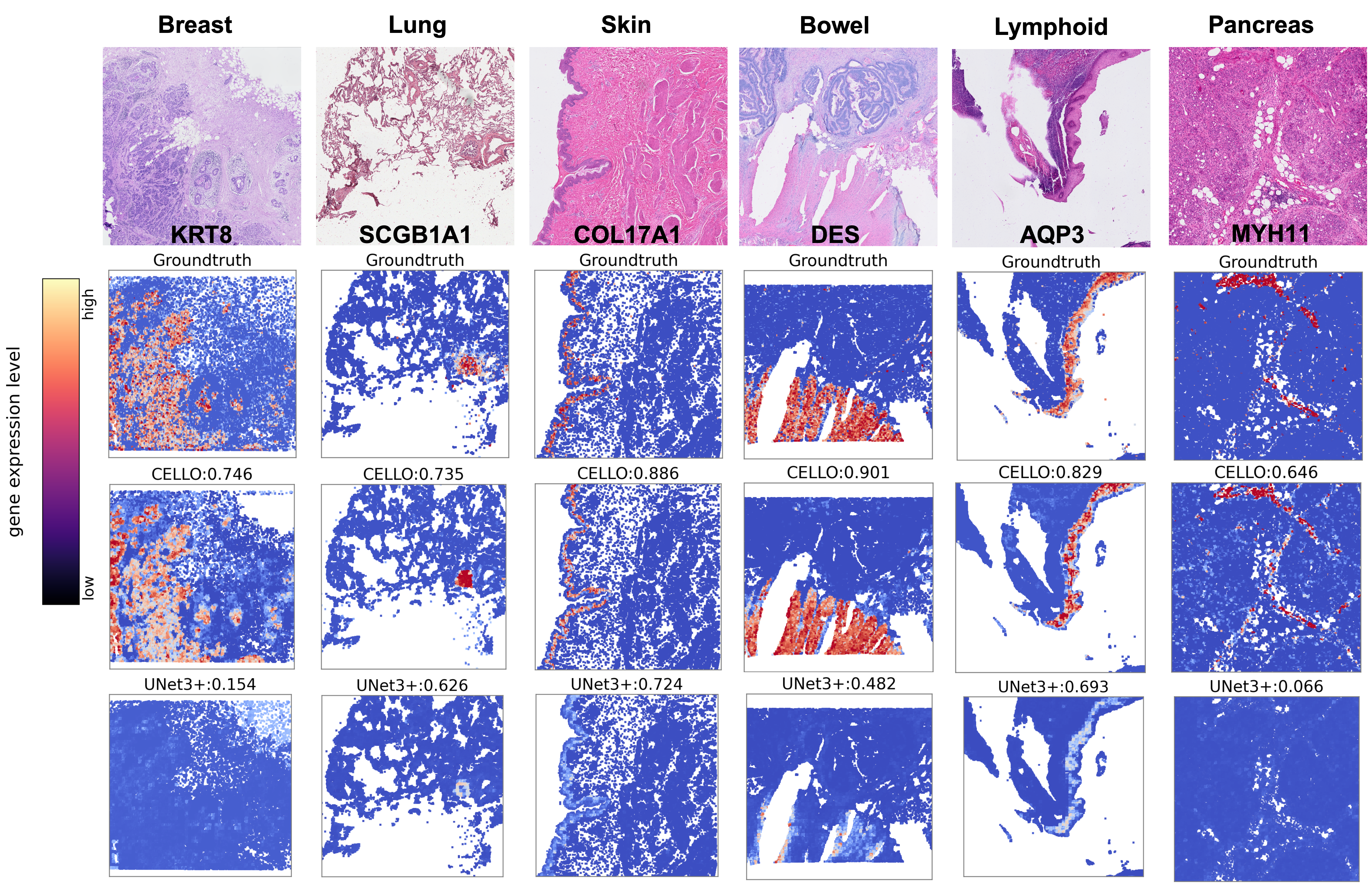}
  \caption{\textbf{Visualization of spatial gene expression predicted from H\&E.} Rows show, from top to bottom: H\&E image, measured gene expression, CELLO predictions, and UNet3+ predictions. Compared with UNet3+, CELLO produced substantially higher correlations with ground-truth gene expression.}

  \label{fig:visualize}
\end{figure}

Figure~\ref{fig:visualize} visualizes H\&E images, ground-truth expression, and model predictions across six representative test tissues using genes KRT8, SCGB1A1, COL17A1, DES, AQP3, and MYH11~\cite{moll1982catalog,rawlins2009role,matsumura2016hair,hara2012chemokine,paulin2004desmin,owens2004molecular}.  Across these heterogeneous morphologies, CELLO closely recovers the spatial localization of measured gene expression.

Compared with UNet3+, CELLO produces sharper and more spatially faithful expression maps, preserving both high-expression domains and low-expression background regions. Quantitatively, CELLO achieves substantially higher Pearson correlations with ground truth for all six markers (0.746, 0.735, 0.886, 0.901, 0.829, and 0.646), whereas UNet3+ shows weaker localization and lower correlations, particularly for KRT8, DES, and MYH11. These qualitative results demonstrate that CELLO can infer cell-level, spatially resolved transcriptional programs directly from H\&E morphology and better capture local tissue context than mask-based baseline prediction.

\section{Conclusion}
\label{main:conclusion}

In this work, we propose CELLO, a scalable end-to-end framework that predicts single-cell spatial transcriptomics from H\&E-stained histology images. By performing a single pathology foundation model forward pass per image and leveraging grid sampling to simultaneously extract features for all cells, CELLO eliminates the per-cell inference bottleneck. The distance-decay cross-attention module encodes biologically motivated spatial priors, enabling each cell to attend preferentially to its local morphological neighborhood while still capturing broader tissue context. Extensive experiments on 52 Xenium whole-slide images spanning 12 organs demonstrate that CELLO achieves state-of-the-art prediction accuracy in both ID and OOD settings while being up to 7.1$\times$ faster at inference. We hope CELLO establishes a strong foundation for scalable, single-cell resolution morphological-to-molecular translation.

\bibliography{cite}
\bibliographystyle{unsrt}

\clearpage
\appendix

\section{Additional Related Work}
\subsection{Context-Aware Modeling in Computational Biology}
Biological processes are inherently context-dependent: a protein's function
varies across tissue microenvironments, a gene's expression is shaped by
neighboring cells, and a molecule's binding affinity shifts with its
interaction partners.
Explicitly encoding such contextual priors into model design has proven
effective across a wide range of biological tasks.
PINNACLE~\cite{li2024contextual} introduces a geometric deep learning framework that
generates context-aware protein representations by modeling protein-protein
interactions conditioned on diverse biological contexts, substantially improving embedding quality. AlphaFold~\cite{jumper2021highly} leverages multiple sequence alignments (MSAs) to retrieve evolutionary co-variation signals from homologous sequences, providing rich phylogenetic context that is critical to its breakthrough accuracy in
protein structure prediction. In spatial transcriptomics, SToFM~\cite{zhao2025stofm} constructs multi-scale contextual representations, from subcellular to tissue levels, to pretrain a more expressive foundation model. COMMOT~\cite{cang2023screening} explicitly models the spatial context of receptor-ligand co-localization to infer intercellular communication patterns. A common thread across these advances is that incorporating biologically motivated inductive biases, rather than relying solely on increased model parameters, can yield substantial performance gains.
This perspective suggests that \emph{context} constitutes a complementary and
largely under-explored scaling axis: while most recent progress in biological
AI has been driven by scaling model parameters and training data, enriching
the contextual information available to a model may offer an equally
impactful path toward improved performance. CELLO embraces this philosophy by equipping each cell query with explicit spatial position context and surrounding tissue morphology.

\section{Dataset Details}
\label{app:dataset}
\paragraph{Data source, registration and preprocessing.}
All 52 H\&E--Xenium pairs are public samples from HEST-1k~\cite{jaume2024hest}; we generated no new tissue or sequencing data. For every Xenium sample, the H\&E image and the transcript readout come from the same tissue section: the section is stained with H\&E and re-imaged after in situ decoding, and 10x Genomics provides an affine transform between the two images. HEST-1k uses this alignment, re-estimates the pixel size of each slide and converts all samples to one coordinate convention. We use the HEST-1k registration and quality control as provided and perform no additional alignment. Each cell is represented by its segmentation centroid in whole-slide pixel coordinates. Each WSI is tiled into non-overlapping $224\times224$ patches on a grid anchored at the slide origin, and a patch is kept if it lies entirely inside the slide and contains at least one cell centroid, giving 520,402 patches and 9.53M cells. Expression is $\log(1+x)$ of counts normalized to 100 per cell. The final split contains 36 training samples, 4 validation samples, and 12 test samples, with the detailed organ, health-condition, and sample-ID composition reported in Table~\ref{tab:dataset_split_details}.
\begin{table*}[!h]
    \centering
    \small
    \caption{Dataset split composition of the 10X-Xenium-52. Sample IDs are grouped by data split, organ type, and health condition.}
    \label{tab:dataset_split_details}
    \begin{tabularx}{\textwidth}{lllrX}
    \toprule
    Split & Organ & Health condition & \# samples & Sample IDs \\
    \midrule
    Train & Bowel & Cancer & 3 & \texttt{TENX111}, \texttt{TENX139}, \texttt{TENX147} \\
    Train & Bowel & Healthy & 1 & \texttt{TENX114} \\
    Train & Breast & Cancer & 4 & \texttt{NCBI783}, \texttt{TENX96}, \texttt{TENX97}, \texttt{TENX98} \\
    Train & Kidney & Cancer & 1 & \texttt{TENX105} \\
    Train & Liver & Healthy & 1 & \texttt{TENX121} \\
    Train & Lung & Cancer & 1 & \texttt{TENX141} \\
    Train & Lung & Diseased & 15 & \texttt{NCBI856}, \texttt{NCBI857}, \texttt{NCBI858}, \texttt{NCBI859}, \texttt{NCBI860}, \texttt{NCBI861}, \texttt{NCBI864}, \texttt{NCBI865}, \texttt{NCBI866}, \texttt{NCBI867}, \texttt{NCBI870}, \texttt{NCBI873}, \texttt{NCBI880}, \texttt{NCBI881}, \texttt{NCBI882} \\
    Train & Lung & Healthy & 3 & \texttt{NCBI875}, \texttt{NCBI876}, \texttt{NCBI884} \\
    Train & Lymphoid & Cancer & 1 & \texttt{TENX134} \\
    Train & Lymphoid & Healthy & 1 & \texttt{TENX133} \\
    Train & Pancreas & Cancer & 2 & \texttt{TENX126}, \texttt{TENX140} \\
    Train & Skin & Cancer & 2 & \texttt{TENX115}, \texttt{TENX117} \\
    Train & Skin & Healthy & 1 & \texttt{TENX123} \\
    \midrule
    Val & Bowel & Cancer & 1 & \texttt{TENX149} \\
    Val & Breast & Cancer & 1 & \texttt{NCBI785} \\
    Val & Lung & Cancer & 1 & \texttt{TENX118} \\
    Val & Lung & Healthy & 1 & \texttt{NCBI883} \\
    \midrule
    Test & Bone & Healthy & 1 & \texttt{TENX132} \\
    Test & Bowel & Cancer & 1 & \texttt{TENX148} \\
    Test & Brain & Cancer & 1 & \texttt{TENX138} \\
    Test & Breast & Cancer & 1 & \texttt{NCBI784} \\
    Test & Heart & Healthy & 1 & \texttt{TENX119} \\
    Test & Kidney & Healthy & 1 & \texttt{TENX106} \\
    Test & Liver & Cancer & 1 & \texttt{TENX120} \\
    Test & Lung & Healthy & 1 & \texttt{NCBI879} \\
    Test & Lymphoid & Diseased & 1 & \texttt{TENX124} \\
    Test & Ovary & Cancer & 1 & \texttt{TENX142} \\
    Test & Pancreas & Cancer & 1 & \texttt{TENX116} \\
    Test & Skin & Healthy & 1 & \texttt{TENX122} \\
    \bottomrule
    \end{tabularx}
\end{table*}

\begin{figure}[!htbp]
  \centering
  \includegraphics[width=\linewidth]{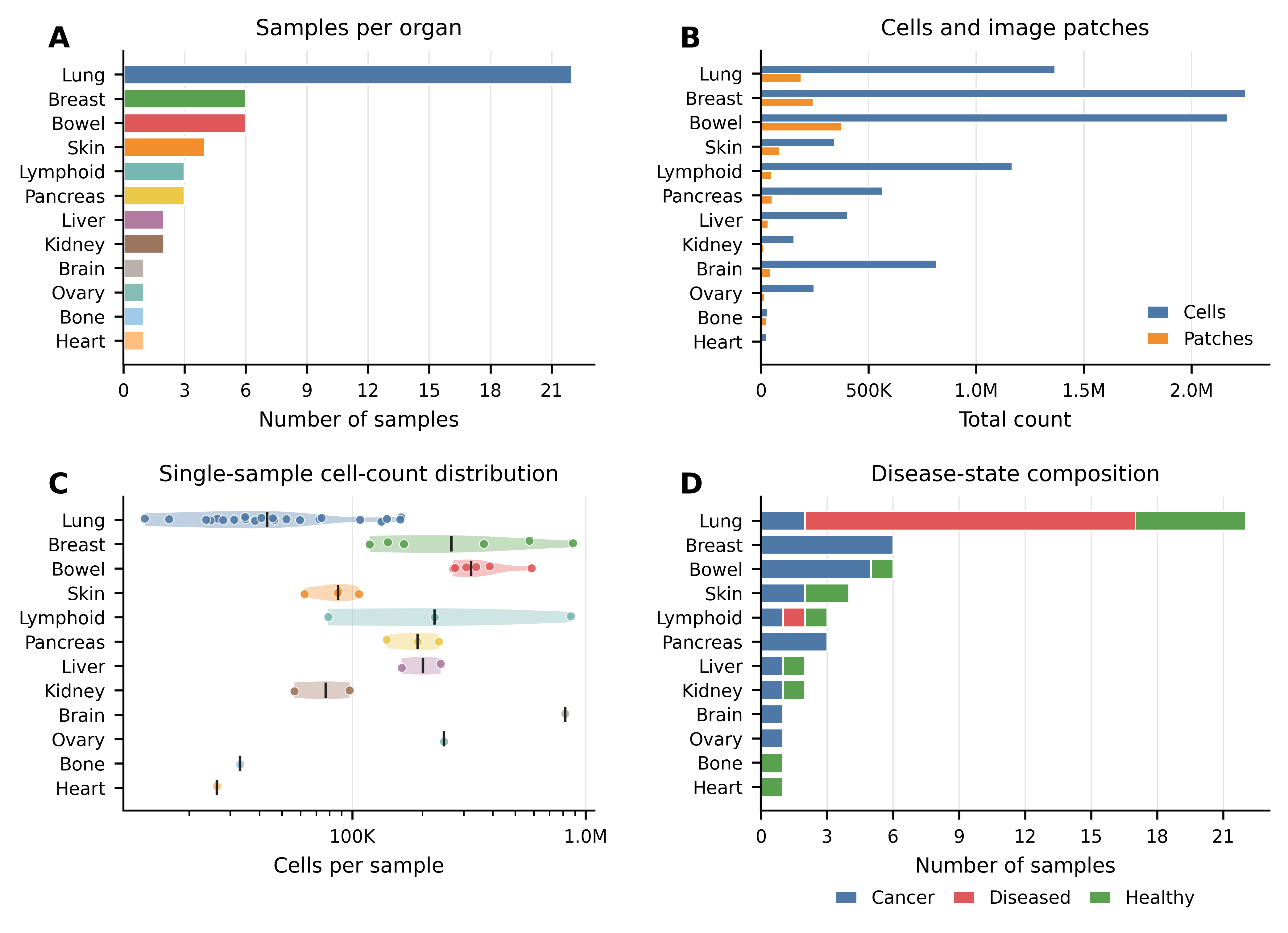}
  \caption{\textbf{Distribution of the 10X-Xenium-52.}
\textbf{(A)} Number of samples per organ. 
\textbf{(B)} Total number of cells and image patches across organs. 
\textbf{(C)} Per-sample cell-count distribution on a log scale, with points denoting individual samples and vertical bars denoting organ-level medians. 
\textbf{(D)} Disease-state composition across organs, including cancer, diseased, and healthy samples.}

  \label{fig:statistic}
\end{figure}

The curated dataset spans 12 organ types with substantial heterogeneity in both tissue origin and sample scale. As shown in Fig.~\ref{fig:statistic}A, Lung is the most heavily represented organ with 22 samples, followed by Breast and Bowel (6 each), Skin (4), Lymphoid and Pancreas (3 each), Liver and Kidney (2 each). Brain, Ovary, Bone and Heart each contribute only 1 sample, yielding a long-tailed distribution across 52 total samples. This imbalance creates a realistic yet challenging benchmark: well-represented organs support in-distribution learning, while single-sample organs provide a stringent test of cross-organ generalization.

Importantly, total cell and patch counts are not simply proportional to sample numbers (Fig.~\ref{fig:statistic}B). Although Lung has the most samples, Breast contributes the largest aggregate cell count, followed by Bowel and Lung. Lymphoid and Kidney also contain substantial cell populations despite having only 3 and 2 samples each, indicating pronounced variation in tissue density and slide coverage.

At the single-sample level (Fig.~\ref{fig:statistic}C), cell counts span more than one order of magnitude—from roughly 20–40 K (Heart, Bone) to 1M (individual Lymphoid outliers). Lung samples are numerous but relatively moderate in per-slide size, whereas Breast, Bowel and Lymphoid include considerably larger individual slides. Brain is especially notable: a single sample containing about 1M cells, making it an important out-of-distribution stress case that combines high cell count with minimal training support.

Finally, the disease-state composition is also heavily imbalanced (Fig.~\ref{fig:statistic}D). Lung and Lymphoid contain all three categories. Bowel, Skin, Liver and Kidney mix cancer and healthy tissue while Bone and Heart are exclusively healthy, and Pancreas and Ovary are exclusively cancer. Together, these distributions underscore that the benchmark evaluates not only prediction accuracy but also robustness to organ imbalance, slide-scale variation and heterogeneous disease contexts.

\paragraph{Gene panels.}
Xenium panels differ between samples. The model predicts a union vocabulary of 1,915 features, and for each sample the loss and all metrics use only the genes measured in that sample. Excluding Xenium negative-control and antisense probes, a panel contains 280--480 genes (Table~\ref{tab:panel_size}). Table~\ref{tab:per_sample} lists every sample.

\begin{table}[!htbp]
\centering\small
\caption{Measured genes per sample, excluding Xenium control probes.}
\label{tab:panel_size}
\begin{tabular}{lrrrrr}
\toprule
Split & Samples & Mean & Min & Median & Max \\
\midrule
Train & 36 & 365 & 280 & 343 & 480 \\
Val   & 4  & 364 & 313 & 360 & 422 \\
Test  & 12 & 414 & 313 & 400 & 480 \\
\bottomrule
\end{tabular}
\end{table}

\begin{table*}[!htbp]
\centering\scriptsize
\caption{Per-sample statistics of the 10X-Xenium-52 (genes exclude Xenium control probes).}
\label{tab:per_sample}
\begin{tabular}{lllcrrr}
\toprule
Split & Organ & Condition & Sample & Genes & Cells & Patches \\
\midrule
    Train & Bowel & Cancer & \texttt{TENX111} & 425 & 587{,}016 & 13{,}816 \\
    Train & Bowel & Cancer & \texttt{TENX139} & 480 & 388{,}015 & 8{,}858 \\
    Train & Bowel & Cancer & \texttt{TENX147} & 422 & 275{,}875 & 9{,}997 \\
    Train & Bowel & Healthy & \texttt{TENX114} & 325 & 270{,}788 & 37{,}902 \\
    Train & Breast & Cancer & \texttt{NCBI783} & 280 & 141{,}829 & 8{,}177 \\
    Train & Breast & Cancer & \texttt{TENX96} & 380 & 365{,}592 & 24{,}657 \\
    Train & Breast & Cancer & \texttt{TENX97} & 380 & 574{,}426 & 33{,}795 \\
    Train & Breast & Cancer & \texttt{TENX98} & 280 & 882{,}692 & 85{,}887 \\
    Train & Kidney & Cancer & \texttt{TENX105} & 377 & 56{,}504 & 3{,}461 \\
    Train & Liver & Healthy & \texttt{TENX121} & 377 & 239{,}197 & 15{,}332 \\
    Train & Lung & Cancer & \texttt{TENX141} & 480 & 160{,}448 & 8{,}073 \\
    Train & Lung & Diseased & \texttt{NCBI856} & 343 & 46{,}277 & 4{,}134 \\
    Train & Lung & Diseased & \texttt{NCBI857} & 343 & 59{,}609 & 5{,}018 \\
    Train & Lung & Diseased & \texttt{NCBI858} & 343 & 16{,}429 & 2{,}594 \\
    Train & Lung & Diseased & \texttt{NCBI859} & 342 & 26{,}389 & 3{,}193 \\
    Train & Lung & Diseased & \texttt{NCBI860} & 343 & 34{,}807 & 2{,}811 \\
    Train & Lung & Diseased & \texttt{NCBI861} & 343 & 45{,}650 & 3{,}083 \\
    Train & Lung & Diseased & \texttt{NCBI864} & 343 & 107{,}973 & 8{,}581 \\
    Train & Lung & Diseased & \texttt{NCBI865} & 343 & 133{,}193 & 9{,}568 \\
    Train & Lung & Diseased & \texttt{NCBI866} & 343 & 72{,}153 & 4{,}950 \\
    Train & Lung & Diseased & \texttt{NCBI867} & 343 & 140{,}660 & 10{,}983 \\
    Train & Lung & Diseased & \texttt{NCBI870} & 343 & 24{,}643 & 2{,}342 \\
    Train & Lung & Diseased & \texttt{NCBI873} & 343 & 27{,}971 & 3{,}597 \\
    Train & Lung & Diseased & \texttt{NCBI880} & 343 & 38{,}285 & 3{,}242 \\
    Train & Lung & Diseased & \texttt{NCBI881} & 343 & 40{,}775 & 3{,}488 \\
    Train & Lung & Diseased & \texttt{NCBI882} & 343 & 52{,}160 & 4{,}417 \\
    Train & Lung & Healthy & \texttt{NCBI875} & 343 & 31{,}189 & 2{,}365 \\
    Train & Lung & Healthy & \texttt{NCBI876} & 343 & 12{,}915 & 1{,}421 \\
    Train & Lung & Healthy & \texttt{NCBI884} & 343 & 73{,}783 & 5{,}151 \\
    Train & Lymphoid & Cancer & \texttt{TENX134} & 477 & 225{,}339 & 4{,}392 \\
    Train & Lymphoid & Healthy & \texttt{TENX133} & 477 & 78{,}904 & 3{,}796 \\
    Train & Pancreas & Cancer & \texttt{TENX126} & 377 & 140{,}194 & 5{,}099 \\
    Train & Pancreas & Cancer & \texttt{TENX140} & 380 & 234{,}859 & 9{,}211 \\
    Train & Skin & Cancer & \texttt{TENX115} & 282 & 106{,}791 & 4{,}297 \\
    Train & Skin & Cancer & \texttt{TENX117} & 382 & 87{,}499 & 9{,}457 \\
    Train & Skin & Healthy & \texttt{TENX123} & 377 & 62{,}469 & 14{,}630 \\
    \midrule
    Val & Bowel & Cancer & \texttt{TENX149} & 422 & 307{,}385 & 9{,}420 \\
    Val & Breast & Cancer & \texttt{NCBI785} & 313 & 166{,}364 & 6{,}280 \\
    Val & Lung & Cancer & \texttt{TENX118} & 377 & 161{,}894 & 4{,}883 \\
    Val & Lung & Healthy & \texttt{NCBI883} & 343 & 34{,}674 & 2{,}829 \\
    \midrule
    Test & Bone & Healthy & \texttt{TENX132} & 477 & 33{,}005 & 8{,}096 \\
    Test & Bowel & Cancer & \texttt{TENX148} & 422 & 339{,}683 & 9{,}421 \\
    Test & Brain & Cancer & \texttt{TENX138} & 480 & 816{,}574 & 23{,}927 \\
    Test & Breast & Cancer & \texttt{NCBI784} & 313 & 118{,}691 & 4{,}652 \\
    Test & Heart & Healthy & \texttt{TENX119} & 377 & 26{,}328 & 4{,}039 \\
    Test & Kidney & Healthy & \texttt{TENX106} & 377 & 97{,}512 & 5{,}063 \\
    Test & Liver & Cancer & \texttt{TENX120} & 474 & 162{,}626 & 7{,}051 \\
    Test & Lung & Healthy & \texttt{NCBI879} & 343 & 23{,}632 & 2{,}292 \\
    Test & Lymphoid & Diseased & \texttt{TENX124} & 377 & 862{,}385 & 15{,}572 \\
    Test & Ovary & Cancer & \texttt{TENX142} & 480 & 247{,}454 & 7{,}468 \\
    Test & Pancreas & Cancer & \texttt{TENX116} & 474 & 190{,}961 & 19{,}584 \\
    Test & Skin & Healthy & \texttt{TENX122} & 377 & 86{,}381 & 8{,}050 \\
\bottomrule
\end{tabular}
\end{table*}

\section{Training and Inference Parameter Details}
\label{app:training_details}

\subsection{Implementation Details of CELLO}
We train our model with 4 NVIDIA H100 GPUs using PyTorch's DistributedDataParallel (DDP) framework. The per-GPU batch size is set to 128, yielding an effective global batch size of 512. Mixed-precision training (FP16) is enabled to accelerate computation and reduce memory consumption, with gradient scaling applied for numerical stability. Gradients are clipped to a maximum norm of 1.0 to prevent training instability.

The model is optimized using AdamW with a cosine annealing learning rate schedule ($\eta_{\min}=10^{-8}$) over 200 epochs. We adopt a two-group parameter strategy to balance the pretrained backbone and the newly initialized task-specific heads: the backbone uses a base learning rate of $5\times 10^{-6}$ with a weight decay of 0.01, while the heads use a learning rate of $2\times 10^{-4}$ with a weight decay of 0.05. To preserve the pretrained representations during early training, we employ a progressive unfreezing schedule: it is fully frozen for the first 10\% of epochs, gradually unfrozen with a linearly increasing learning rate (up to $2\times 10^{-5}$) over the next 30\% of epochs, and fully fine-tuned thereafter.

The overall training objective combines a cell-level gene regression loss and a patch-level gene prediction loss, weighted by 1.0 and 0.5, respectively. Validation is performed every 5 epochs using whole-slide-image (WSI) sliding-window inference with a stride equal to the patch size (224), and the model achieving the highest gene-level Pearson correlation coefficient (PCC) on the validation set is retained as the best checkpoint. Random seeds are fixed (seed $=42$) for reproducibility.

For WSI-level evaluation, we used non-overlapping \(224 \times 224\) patches with stride 224, matching the input resolution used during training. Each cell was assigned to the patch containing its spatial coordinate, and predictions were generated using only the visual tokens within that patch.

\subsection{Implementation Details of DeepSpot2Cell}

For DeepSpot2Cell, we follow the original design of the method and keep the pathology foundation model frozen. This choice is mainly practical: DeepSpot2Cell extracts PFM features independently for each segmented cell crop, and fine-tuning the PFM would require backpropagating through a separate large-encoder forward pass for every cell, resulting in prohibitive training cost on million-cell WSIs. In contrast, CELLO amortizes the PFM computation at the patch level and therefore can be trained end-to-end with progressive backbone unfreezing. We note that this difference may partly contribute to the performance gap, since CELLO benefits from task-specific representation adaptation. However, it also reflects an inherent scalability limitation of per-cell-crop PFM baselines: making them end-to-end trainable is substantially more expensive than patch-level amortized training.

\section{Additional Gene Panel Performance}
\label{app:full_gene_results}
In the main experiments, we report performance on most commonly used and clinically significant gene subsets: highly variable genes (denoted as HVG). Here we report another two selected subsets: most predictive genes (MPG) and spatially variable genes (SVG). As shown in Table~\ref{tab:main_mpg_svg}, CELLO also substantially outperforms UNet3+ and DeepSpot2Cell in both ID and OOD settings. In the ID setting, CELLO improves over the strongest baseline, DeepSpot2Cell, by 70.6\% on MPG PCC-50 and 68.8\% on SVG PCC-50. This advantage becomes even larger in the OOD setting, where CELLO achieves a 117.9\% relative improvement on MPG PCC-50 and 107.1\% on SVG PCC-50, indicating stronger robustness under unseen tissue and disease conditions. In addition, the comparison with \emph{Grid}, an ablated version of CELLO without the distance-decay cross-attention module, further validates the importance of contextual refinement.

Moreover, practical single-cell spatial transcriptomics prediction ultimately requires recovering the broader measured gene panel, including genes with weaker morphology-associated signals and higher sparsity. Therefore, we additionally evaluate all methods using the full Xenium gene panel to assess whether the performance gains of CELLO remain robust beyond curated gene subsets.

\begin{figure}
  \centering
  \includegraphics[width=0.85\linewidth]{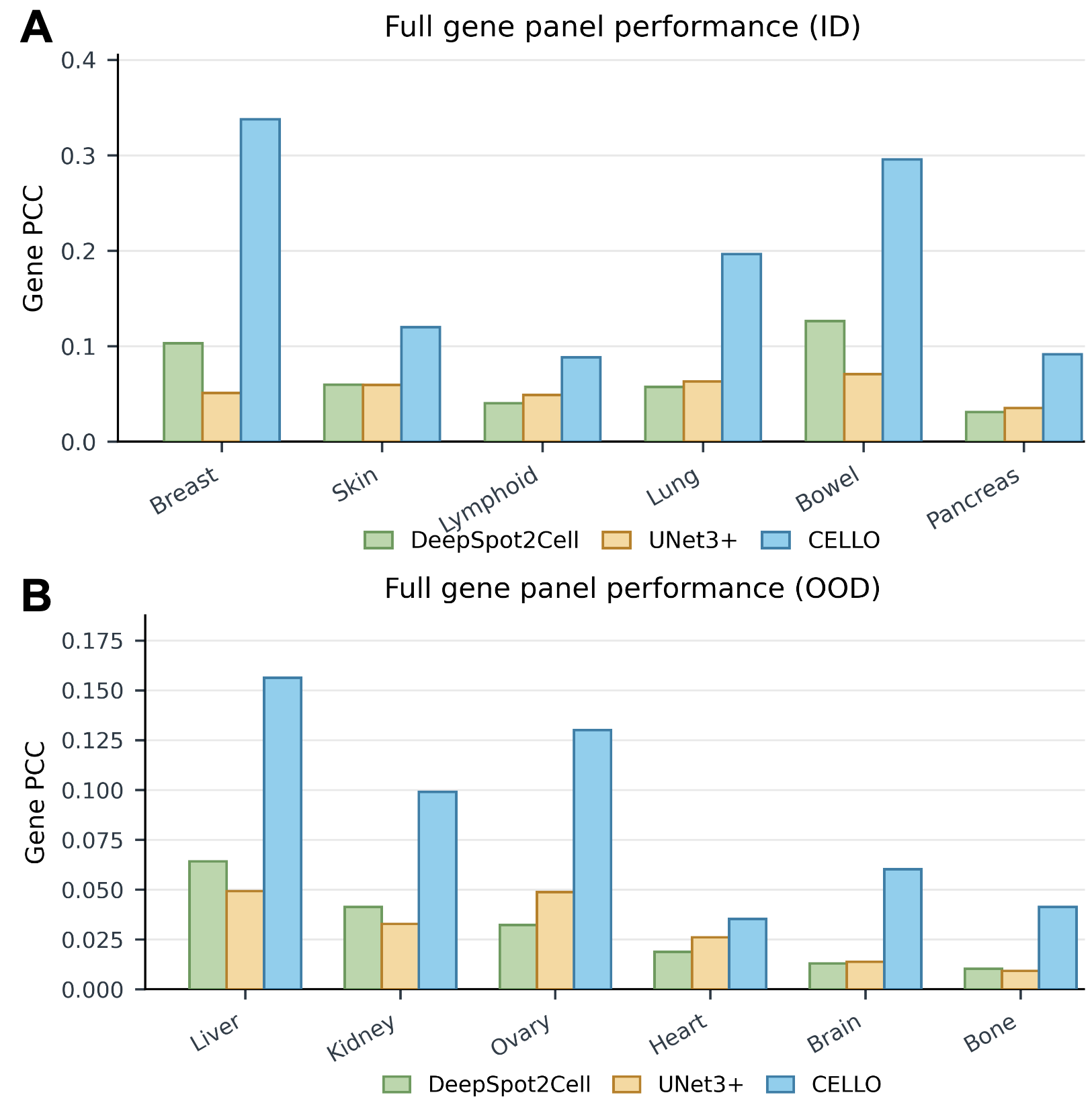}
  \caption{\textbf{Full gene panel performance comparison between DeepSpot2Cell, UNet3+ and CELLO.}
\textbf{(A)} Full gene panel performance of ID setting. 
\textbf{(B)} Full gene panel performance of OOD setting.}
  \label{fig:full_panel}
\end{figure}

As shown in Fig~\ref{fig:full_panel}, CELLO consistently achieves the highest gene-level PCC across both in-distribution (ID) and out-of-distribution (OOD) settings. In the ID setting, CELLO substantially outperforms DeepSpot2Cell and UNet3+ across all six tissues, with particularly large gains in Breast, Bowel, and Lung. This indicates that CELLO’s patch-level feature extraction and context-aware cell representation remain effective even when evaluated over the full, noisier gene panel. In the OOD setting, overall correlations are lower, reflecting the increased difficulty of generalizing to unseen tissues and disease contexts. Nevertheless, CELLO maintains a clear advantage across all organs, especially in Liver, Ovary, and Kidney, demonstrating stronger cross-tissue robustness. These results further support that CELLO improves not only selected high-signal genes but also broader panel-wide gene expression prediction.

\begin{table}[!htbp]
    \centering
    \caption{\textbf{MPG and SVG results on the 10X-Xenium-52 benchmark.} ID tissues are shown on the left and OOD tissues on the right. Higher PCC values are better. The best results within each tissue are highlighted in \textbf{bold}.}
    \label{tab:main_mpg_svg}
    \tiny
    \setlength{\tabcolsep}{1.8pt}
    \resizebox{\textwidth}{!}{%
    \begin{tabular}{llcccccc@{\hspace{3pt}}!{\vrule width 0.35pt}@{\hspace{3pt}}llcccccc}
    \toprule
    \multicolumn{8}{c}{\textbf{In-distribution (ID)}} &
    \multicolumn{8}{c}{\textbf{Out-of-distribution (OOD)}} \\
    \cmidrule(lr){1-8} \cmidrule(lr){9-16}
    Tissue & Model & M-10 & M-50 & M-200 & S-10 & S-50 & S-200 &
    Tissue & Model & M-10 & M-50 & M-200 & S-10 & S-50 & S-200 \\
    \midrule

    \multirow{4}{*}{Lung} & UNet3+        & 0.4530 & 0.2470 & 0.1322 & 0.4530 & 0.2470 & 0.1225
    & \multirow{4}{*}{Kidney} & UNet3+        & 0.2065 & 0.1356 & 0.0697 & 0.2065 & 0.1348 & 0.0479 \\
    & DeepSpot2Cell & 0.4232 & 0.2580 & 0.1387 & 0.4232 & 0.2580 & 0.1279
    & & DeepSpot2Cell & 0.3617 & 0.1975 & 0.0871 & 0.3617 & 0.1970 & 0.0717 \\
    & Grid          & 0.5492 & 0.3956 & 0.2390 & 0.5492 & 0.3956 & 0.2234
    & & Grid          & 0.4879 & 0.3431 & 0.1730 & 0.4879 & 0.3431 & 0.1483 \\
    & \textbf{CELLO} & \textbf{0.5538} & \textbf{0.4060} & \textbf{0.2450} & \textbf{0.5538} & \textbf{0.4060} & \textbf{0.2286}
    & & \textbf{CELLO} & \textbf{0.5648} & \textbf{0.3838} & \textbf{0.2085} & \textbf{0.5648} & \textbf{0.3753} & \textbf{0.1610} \\
    \cmidrule(lr){1-8} \cmidrule(lr){9-16}

    \multirow{4}{*}{Breast} & UNet3+        & 0.3126 & 0.2138 & 0.1048 & 0.3126 & 0.2138 & 0.0778
    & \multirow{4}{*}{Heart} & UNet3+        & 0.2166 & 0.1255 & 0.0600 & 0.2166 & 0.1255 & 0.0451 \\
    & DeepSpot2Cell & 0.5353 & 0.3772 & 0.1793 & 0.5353 & 0.3772 & 0.1656
    & & DeepSpot2Cell & 0.2115 & 0.1154 & 0.0482 & 0.2115 & 0.1153 & 0.0382 \\
    & Grid          & 0.6616 & 0.5407 & 0.3229 & 0.6616 & 0.5407 & 0.2995
    & & Grid          & 0.3490 & 0.2030 & 0.0971 & 0.3490 & 0.2029 & 0.0843 \\
    & \textbf{CELLO} & \textbf{0.6995} & \textbf{0.5681} & \textbf{0.3473} & \textbf{0.6995} & \textbf{0.5681} & \textbf{0.3217}
    & & \textbf{CELLO} & \textbf{0.4499} & \textbf{0.2907} & \textbf{0.1197} & \textbf{0.4499} & \textbf{0.2896} & \textbf{0.0848} \\
    \cmidrule(lr){1-8} \cmidrule(lr){9-16}

    \multirow{4}{*}{Bowel} & UNet3+        & 0.4071 & 0.2904 & 0.1739 & 0.4071 & 0.2902 & 0.1245
    & \multirow{4}{*}{Liver} & UNet3+        & 0.3957 & 0.2719 & 0.1362 & 0.3957 & 0.2712 & 0.0938 \\
    & DeepSpot2Cell & 0.6592 & 0.5041 & 0.2973 & 0.6592 & 0.5041 & 0.2608
    & & DeepSpot2Cell & 0.4640 & 0.3534 & 0.1773 & 0.4640 & 0.3534 & 0.1316 \\
    & Grid          & 0.6911 & 0.5463 & 0.3592 & 0.6911 & 0.5463 & 0.3273
    & & Grid          & 0.5301 & 0.4064 & 0.2292 & 0.5301 & 0.4064 & 0.1818 \\
    & \textbf{CELLO} & \textbf{0.7944} & \textbf{0.7268} & \textbf{0.5703} & \textbf{0.7944} & \textbf{0.7241} & \textbf{0.4892}
    & & \textbf{CELLO} & \textbf{0.6840} & \textbf{0.6118} & \textbf{0.3995} & \textbf{0.6840} & \textbf{0.5881} & \textbf{0.2234} \\
    \cmidrule(lr){1-8} \cmidrule(lr){9-16}

    \multirow{4}{*}{Skin} & UNet3+        & 0.4365 & 0.2639 & 0.1185 & 0.4365 & 0.2639 & 0.1121
    & \multirow{4}{*}{Bone} & UNet3+        & 0.1109 & 0.0705 & 0.0349 & 0.1109 & 0.0667 & 0.0134 \\
    & DeepSpot2Cell & 0.5781 & 0.3310 & 0.1306 & 0.5781 & 0.3310 & 0.1253
    & & DeepSpot2Cell & 0.1435 & 0.0818 & 0.0314 & 0.1435 & 0.0792 & 0.0206 \\
    & Grid          & 0.6613 & 0.4530 & 0.2096 & 0.6613 & 0.4530 & 0.2019
    & & Grid          & 0.1838 & 0.1112 & 0.0498 & 0.1838 & 0.1093 & 0.0360 \\
    & \textbf{CELLO} & \textbf{0.7921} & \textbf{0.6264} & \textbf{0.3092} & \textbf{0.7921} & \textbf{0.6264} & \textbf{0.2836}
    & & \textbf{CELLO} & \textbf{0.2872} & \textbf{0.2189} & \textbf{0.1068} & \textbf{0.2812} & \textbf{0.1842} & \textbf{0.0566} \\
    \cmidrule(lr){1-8} \cmidrule(lr){9-16}

    \multirow{4}{*}{Pancreas} & UNet3+        & 0.2761 & 0.1760 & 0.0903 & 0.2761 & 0.1755 & 0.0586
    & \multirow{4}{*}{Brain} & UNet3+        & 0.2056 & 0.0940 & 0.0404 & 0.2056 & 0.0933 & 0.0267 \\
    & DeepSpot2Cell & 0.3190 & 0.2125 & 0.0848 & 0.3190 & 0.2086 & 0.0598
    & & DeepSpot2Cell & 0.2054 & 0.1076 & 0.0388 & 0.2054 & 0.1070 & 0.0273 \\
    & Grid          & 0.3388 & 0.2272 & 0.1183 & 0.3379 & 0.2164 & 0.0764
    & & Grid          & 0.4054 & 0.2046 & 0.0837 & \textbf{0.4254} & 0.2036 & 0.0677 \\
    & \textbf{CELLO} & \textbf{0.5311} & \textbf{0.4030} & \textbf{0.2419} & \textbf{0.5095} & \textbf{0.3726} & \textbf{0.1190}
    & & \textbf{CELLO} & \textbf{0.4145} & \textbf{0.2949} & \textbf{0.1539} & 0.3906 & \textbf{0.2511} & \textbf{0.0950} \\
    \cmidrule(lr){1-8} \cmidrule(lr){9-16}

    \multirow{4}{*}{Lymphoid} & UNet3+        & 0.3948 & 0.2567 & 0.1181 & 0.3948 & 0.2567 & 0.1111
    & \multirow{4}{*}{Ovary} & UNet3+        & 0.4089 & 0.2358 & 0.1096 & 0.4089 & 0.2358 & 0.0952 \\
    & DeepSpot2Cell & 0.3806 & 0.2493 & 0.1034 & 0.3806 & 0.2493 & 0.0943
    & & DeepSpot2Cell & 0.3981 & 0.2117 & 0.0847 & 0.3981 & 0.2106 & 0.0666 \\
    & Grid          & 0.5010 & 0.3757 & 0.1795 & 0.5110 & 0.3708 & 0.1649
    & & Grid          & 0.5184 & 0.3189 & 0.1490 & 0.5184 & 0.3172 & 0.1289 \\
    & \textbf{CELLO} & \textbf{0.7351} & \textbf{0.5661} & \textbf{0.2594} & \textbf{0.7351} & \textbf{0.5577} & \textbf{0.2068}
    & & \textbf{CELLO} & \textbf{0.6421} & \textbf{0.5263} & \textbf{0.3106} & \textbf{0.6406} & \textbf{0.5112} & \textbf{0.2187} \\
    \midrule

    \multirow{4}{*}{Average} & UNet3+        & 0.3800 & 0.2413 & 0.1230 & 0.3800 & 0.2412 & 0.1011
    & \multirow{4}{*}{Average} & UNet3+        & 0.2574 & 0.1556 & 0.0751 & 0.2574 & 0.1546 & 0.0537 \\
    & DeepSpot2Cell & 0.4826 & 0.3220 & 0.1557 & 0.4826 & 0.3214 & 0.1390
    & & DeepSpot2Cell & 0.2974 & 0.1779 & 0.0779 & 0.2974 & 0.1771 & 0.0593 \\
    & Grid          & 0.5672 & 0.4231 & 0.2381 & 0.5687 & 0.4205 & 0.2156
    & & Grid          & 0.4158 & 0.2645 & 0.1303 & 0.4158 & 0.2638 & 0.1078 \\
    & \textbf{CELLO} & \textbf{0.6843} & \textbf{0.5494} & \textbf{0.3289} & \textbf{0.6807} & \textbf{0.5425} & \textbf{0.2748}
    & & \textbf{CELLO} & \textbf{0.5071} & \textbf{0.3877} & \textbf{0.2165} & \textbf{0.5018} & \textbf{0.3666} & \textbf{0.1399} \\

    \bottomrule
    \end{tabular}%
    }
    \vspace{-0.5em}
    \begin{flushleft}
    \tiny M and S denote MPG and SVG, respectively.
    \end{flushleft}
\end{table}

\section{Additional Ablation Study Results}
\label{app:ablation}
In this section, we show additional ablation results of encoder choices and cls token context on both ID and OOD setting. The encoder ablation further identified Virchow2 as the most effective pathology foundation encoder. As shown in Table~\ref{tab:main_ablation_id}, in the ID setting, Virchow2 achieved the best averaged result across all nine metrics. Relative to the strongest encoder H0, Virchow2 improved average MPG PCC-10/50/200 by +19.7\%, +28.9\%, and +35.5\%, HVG PCC-10/50/200 by +13.7\%, +14.7\%, and +11.9\%, and SVG PCC-10/50/200 by +19.2\%, +28.1\%, and +25.3\%. This advantage remained clear in the OOD ablation, where Virchow2 achieved the best average performance in eight of the nine metrics. As shown in Table~\ref{tab:main_ablation_ood}, compared with H0, Virchow2 yielded substantial OOD gains on MPG PCC-10/50/200 +17.4\%, +41.9\%, +62.1\% and SVG PCC-10/50/200 +16.1\%, +34.8\%, +26.7\%, while also improving HVG PCC-10 and PCC-50 by +11.8\% and +11.0\%. These results suggest that Virchow2 provides more transferable morphological representations for spatial gene expression prediction.

In contrast, incorporating the CLS token did not provide consistent benefits. For the strongest encoder, Virchow2, adding the CLS token reduced the simple average over the nine averaged metrics from 0.4793 to 0.4143 in the ID setting, corresponding to a 13.6\% relative decrease. A similar trend was observed in the OOD setting, where the average dropped from 0.3298 to 0.2867, a 13.1\% relative decrease. Although a few isolated entries improved, such as OOD HVG PCC-200, these gains were not systematic and were outweighed by broad degradation across MPG and SVG metrics. This indicates that the global CLS representation contributes limited useful information for this task and may even dilute the spatially localized morphological signals required for accurate gene expression prediction.

\begin{table}[htbp]
    \centering
    \caption{\textbf{Ablation results on the 10X-Xenium-52 ID setting evaluated by MPG, HVG, and SVG metrics.} We compare different pathology foundation models as vision encoders and investigate whether incorporating the \textbf{CLS} token to provide additional global contextual information improves performance. Higher values of PCC-10, PCC-50, and PCC-200 indicate better results. The best results are highlighted in \textbf{bold}.}
    \label{tab:main_ablation_id}
    
    \resizebox{\textwidth}{!}{%
    \begin{tabular}{l ccc ccc ccc}
    \toprule
    \multirow{2}{*}{{Model}} 
    & \multicolumn{3}{c}{{MPG}} 
    & \multicolumn{3}{c}{{HVG}} 
    & \multicolumn{3}{c}{{SVG}} \\
    \cmidrule(lr){2-4} \cmidrule(lr){5-7} \cmidrule(lr){8-10}
    & {PCC-10$\uparrow$} 
    & {PCC-50$\uparrow$} 
    & {PCC-200$\uparrow$} 
     & {PCC-10$\uparrow$} 
    & {PCC-50$\uparrow$} 
    & {PCC-200$\uparrow$} 
     & {PCC-10$\uparrow$} 
    & {PCC-50$\uparrow$} 
    & {PCC-200$\uparrow$}  \\
    \midrule

    \rowcolor{cyan!8}
    \multicolumn{10}{l}{\textit{Lung}} \\
    UNI          & 0.5451 & 0.3969 & 0.2410 & 0.5451 & 0.3753 & 0.1492 & 0.5451 & 0.3969 & 0.2253 \\
    UNI\_cls     & 0.5441 & 0.3931 & 0.2368 & 0.5441 & 0.3714 & 0.1437 & 0.5441 & 0.3931 & 0.2204 \\
    H0           & 0.5533 & 0.4053 & 0.2457 & 0.5533 & 0.3795 & 0.1485 & 0.5533 & 0.4053 & 0.2292 \\
    H0\_cls      & 0.4892 & 0.3734 & 0.2213 & 0.4880 & 0.3427 & 0.1249 & 0.4892 & 0.3734 & 0.1960 \\
    Virchow2     & \textbf{0.5538} & 0.4060 & 0.2450 & 0.5537 & 0.3827 & 0.1492 & \textbf{0.5538} & 0.4060 & 0.2286 \\
    Virchow2\_cls& 0.5528 & \textbf{0.4095} & \textbf{0.2497} & \textbf{0.5538} & \textbf{0.3832} & \textbf{0.1535} & 0.5528 & \textbf{0.4095} & \textbf{0.2325} \\
    \addlinespace[4pt]

    \rowcolor{blue!6}
    \multicolumn{10}{l}{\textit{Breast}} \\
    UNI          & 0.6417 & 0.5315 & 0.3174 & 0.6342 & 0.4946 & 0.2316 & 0.6417 & 0.5315 & 0.2972 \\
    UNI\_cls     & 0.6515 & 0.5316 & 0.3243 & 0.6439 & 0.5023 & 0.2407 & 0.6515 & 0.5316 & 0.3013 \\
    H0           & 0.6065 & 0.4800 & 0.2962 & 0.6065 & 0.4696 & 0.2308 & 0.6065 & 0.4800 & 0.2744 \\
    H0\_cls      & 0.5986 & 0.4619 & 0.2527 & 0.5986 & 0.4581 & 0.1980 & 0.5986 & 0.4619 & 0.2207 \\
    Virchow2     & 0.6995 & \textbf{0.5681} & \textbf{0.3473} & \textbf{0.6736} & \textbf{0.5243} & \textbf{0.2541} & 0.6995 & \textbf{0.5681} & \textbf{0.3217} \\
    Virchow2\_cls& \textbf{0.7008} & 0.5643 & 0.3421 & 0.6717 & 0.5157 & 0.2476 & \textbf{0.7008} & 0.5643 & 0.3193 \\
    \addlinespace[4pt]

    \rowcolor{cyan!12}
    \multicolumn{10}{l}{\textit{Bowel}} \\
    UNI          & 0.6974 & 0.5428 & 0.3612 & 0.6784 & 0.4793 & 0.2148 & 0.6974 & 0.5428 & 0.3341 \\
    UNI\_cls     & 0.7010 & 0.5544 & 0.3680 & 0.6809 & 0.4935 & 0.2195 & 0.7010 & 0.5544 & 0.3383 \\
    H0           & 0.7088 & 0.5545 & 0.3669 & 0.6989 & 0.5113 & 0.2244 & 0.7088 & 0.5545 & 0.3334 \\
    H0\_cls      & 0.6969 & 0.5295 & 0.3344 & 0.6700 & 0.4767 & 0.1820 & 0.6969 & 0.5295 & 0.2761 \\
    Virchow2     & \textbf{0.7944} & \textbf{0.7268} & \textbf{0.5703} & \textbf{0.7732} & \textbf{0.6599} & \textbf{0.3313} & \textbf{0.7944} & \textbf{0.7241} & \textbf{0.4892} \\
    Virchow2\_cls& 0.7116 & 0.5739 & 0.3843 & 0.6946 & 0.5170 & 0.2340 & 0.7116 & 0.5739 & 0.3531 \\
    \addlinespace[4pt]

    \rowcolor{blue!14}
    \multicolumn{10}{l}{\textit{Skin}} \\
    UNI          & 0.6500 & 0.4230 & 0.1974 & 0.6141 & 0.3444 & 0.1214 & 0.6500 & 0.4230 & 0.1901 \\
    UNI\_cls     & 0.6562 & 0.4082 & 0.1828 & 0.6373 & 0.3309 & 0.1142 & 0.6562 & 0.4082 & 0.1768 \\
    H0           & 0.7040 & 0.4990 & 0.2359 & 0.6864 & 0.3901 & \textbf{0.1367} & 0.7040 & 0.4990 & 0.2265 \\
    H0\_cls      & 0.6712 & 0.4326 & 0.1895 & 0.6413 & 0.3313 & 0.1092 & 0.6712 & 0.4326 & 0.1792 \\
    Virchow2     & \textbf{0.7921} & \textbf{0.6264} & \textbf{0.3092} & \textbf{0.7672} & \textbf{0.4503} & 0.1302 & \textbf{0.7921} & \textbf{0.6264} & \textbf{0.2836} \\
    Virchow2\_cls& 0.6888 & 0.4949 & 0.2346 & 0.6765 & 0.3879 & 0.1358 & 0.6888 & 0.4949 & 0.2259 \\
    \addlinespace[4pt]

    \rowcolor{blue!10}
    \multicolumn{10}{l}{\textit{Pancreas}} \\
    UNI          & 0.3372 & 0.2327 & 0.1215 & 0.3247 & 0.2248 & 0.0857 & 0.3372 & 0.2253 & 0.0797 \\
    UNI\_cls     & 0.3382 & 0.2377 & 0.1274 & 0.3260 & 0.2296 & 0.0960 & 0.3364 & 0.2284 & 0.0868 \\
    H0           & 0.3510 & 0.2383 & 0.1237 & 0.3469 & 0.2252 & 0.0864 & 0.3484 & 0.2237 & 0.0797 \\
    H0\_cls      & 0.3640 & 0.2544 & 0.1289 & 0.3553 & 0.2343 & 0.0828 & 0.3637 & 0.2402 & 0.0886 \\
    Virchow2     & \textbf{0.5311} & \textbf{0.4030} & \textbf{0.2419} & \textbf{0.4181} & 0.2640 & 0.0767 & \textbf{0.5095} & \textbf{0.3726} & \textbf{0.1190} \\
    Virchow2\_cls& 0.4169 & 0.2906 & 0.1573 & 0.4101 & \textbf{0.2790} & \textbf{0.1144} & 0.4169 & 0.2807 & 0.1121 \\
    \addlinespace[4pt]

    \rowcolor{blue!6}
    \multicolumn{10}{l}{\textit{Lymphoid}} \\
    UNI          & 0.4747 & 0.3378 & 0.1589 & 0.4570 & 0.2703 & 0.0803 & 0.4747 & 0.3356 & 0.1460 \\
    UNI\_cls     & 0.4657 & 0.3376 & 0.1645 & 0.4469 & 0.2771 & 0.0825 & 0.4657 & 0.3333 & 0.1479 \\
    H0           & 0.5063 & 0.3798 & 0.1886 & 0.4965 & 0.3206 & \textbf{0.0943} & 0.5063 & 0.3778 & 0.1725 \\
    H0\_cls      & 0.5101 & 0.3782 & 0.1765 & 0.4953 & 0.2960 & 0.0816 & 0.5101 & 0.3758 & 0.1610 \\
    Virchow2     & \textbf{0.7351} & \textbf{0.5661} & \textbf{0.2594} & \textbf{0.6667} & \textbf{0.3528} & 0.0891 & \textbf{0.7351} & \textbf{0.5577} & \textbf{0.2068} \\
    Virchow2\_cls& 0.5083 & 0.3734 & 0.1814 & 0.4848 & 0.3019 & 0.0899 & 0.5083 & 0.3715 & 0.1650 \\
    \addlinespace[4pt]

    \rowcolor{gray!15}
    \multicolumn{10}{l}{\textit{Average}} \\
    UNI          & 0.5577 & 0.4108 & 0.2329 & 0.5423 & 0.3648 & 0.1472 & 0.5577 & 0.4092 & 0.2121 \\
    UNI\_cls     & 0.5595 & 0.4104 & 0.2340 & 0.5465 & 0.3675 & 0.1494 & 0.5592 & 0.4082 & 0.2119 \\
    H0           & 0.5717 & 0.4262 & 0.2428 & 0.5648 & 0.3827 & 0.1535 & 0.5712 & 0.4234 & 0.2193 \\
    H0\_cls      & 0.5550 & 0.4050 & 0.2172 & 0.5414 & 0.3565 & 0.1298 & 0.5550 & 0.4022 & 0.1869 \\
    \textbf{Virchow2} & \textbf{0.6843} & \textbf{0.5494} & \textbf{0.3289} & \textbf{0.6421} & \textbf{0.4390} & \textbf{0.1718} & \textbf{0.6807} & \textbf{0.5425} & \textbf{0.2748} \\
    Virchow2\_cls& 0.5965 & 0.4511 & 0.2582 & 0.5819 & 0.3975 & 0.1625 & 0.5967 & 0.4491 & 0.2347 \\
    \bottomrule
    \end{tabular}%
    }
\end{table}

\paragraph{Loss Weight and Distance-Decay Bias}
We trained CELLO with $\lambda\in\{0,0.1,0.25,0.5,1.0\}$ and, at $\lambda=0.5$, without the distance-decay bias. All six runs use the same data, one GPU with 16 patches per batch and 50 epochs, so they are comparable with each other but not with the 200-epoch model of the main tables. Table~\ref{tab:ablation_loss} reports HVG PCC averaged over the ID and OOD test slides. Within this budget the settings differ by less than 0.03, $\lambda=0.5$ scores lowest in both settings, and removing the distance-decay bias does not lower performance relative to $\lambda=0.5$. These single-seed short runs therefore do not show a clear effect of either component.

\begin{table}[!htbp]
\centering\small
\caption{Loss weight $\lambda$ and distance-decay bias (50-epoch runs, HVG PCC).}
\label{tab:ablation_loss}
\begin{tabular}{lcccccc}
\toprule
 & \multicolumn{3}{c}{ID} & \multicolumn{3}{c}{OOD} \\
\cmidrule(lr){2-4}\cmidrule(lr){5-7}
Setting & 10 & 50 & 200 & 10 & 50 & 200 \\
\midrule
$\lambda=0$    & 0.5598 & 0.3882 & 0.1752 & 0.4314 & 0.2574 & 0.0975 \\
$\lambda=0.1$  & 0.5565 & 0.3866 & 0.1732 & 0.4178 & 0.2473 & 0.0933 \\
$\lambda=0.25$ & 0.5533 & 0.3823 & 0.1683 & 0.4147 & 0.2484 & 0.0922 \\
$\lambda=0.5$  & 0.5447 & 0.3750 & 0.1665 & 0.4018 & 0.2349 & 0.0850 \\
$\lambda=1.0$  & 0.5614 & 0.3808 & 0.1692 & 0.4298 & 0.2553 & 0.0950 \\
$\lambda=0.5$, no distance bias & 0.5472 & 0.3822 & 0.1685 & 0.4232 & 0.2468 & 0.0908 \\
\bottomrule
\end{tabular}
\end{table}

\begin{table}[htbp]
    \centering
    \caption{\textbf{Ablation results on the 10X-Xenium-52 OOD setting evaluated by MPG, HVG, and SVG metrics.} We compare different pathology foundation models as vision encoders and investigate whether incorporating the \textbf{CLS} token to provide additional global contextual information improves performance. Higher values of PCC-10, PCC-50, and PCC-200 indicate better results. The best results are highlighted in \textbf{bold}.}
    \label{tab:main_ablation_ood}
    
    \resizebox{\textwidth}{!}{%
    \begin{tabular}{l ccc ccc ccc}
    \toprule
    \multirow{2}{*}{{Model}} 
    & \multicolumn{3}{c}{{MPG}} 
    & \multicolumn{3}{c}{{HVG}} 
    & \multicolumn{3}{c}{{SVG}} \\
    \cmidrule(lr){2-4} \cmidrule(lr){5-7} \cmidrule(lr){8-10}
    & {PCC-10$\uparrow$} 
    & {PCC-50$\uparrow$} 
    & {PCC-200$\uparrow$} 
    & {PCC-10$\uparrow$} 
    & {PCC-50$\uparrow$} 
    & {PCC-200$\uparrow$}
    & {PCC-10$\uparrow$} 
    & {PCC-50$\uparrow$} 
    & {PCC-200$\uparrow$} \\
    \midrule

    \rowcolor{teal!7}
    \multicolumn{10}{l}{\textit{Kidney}} \\
    UNI          & 0.5638 & 0.3600 & 0.1783 & 0.5220 & 0.3229 & 0.1269 & 0.5638 & 0.3600 & 0.1545 \\
    UNI\_cls     & 0.5155 & 0.3419 & 0.1737 & 0.4850 & 0.3070 & 0.1249 & 0.5155 & 0.3419 & 0.1515 \\
    H0           & 0.5041 & 0.3605 & 0.1843 & 0.5041 & 0.3404 & 0.1368 & 0.5041 & 0.3604 & 0.1592 \\
    H0\_cls      & 0.4927 & 0.3275 & 0.1628 & 0.4750 & 0.3045 & 0.1127 & 0.4927 & 0.3275 & 0.1378 \\
    Virchow2     & 0.5648 & 0.3838 & \textbf{0.2085} & 0.5515 & 0.2743 & 0.1078 & 0.5648 & 0.3753 & 0.1610 \\
    Virchow2\_cls& \textbf{0.6180} & \textbf{0.3934} & 0.1908 & \textbf{0.5625} & \textbf{0.3468} & \textbf{0.1369} & \textbf{0.6180} & \textbf{0.3934} & \textbf{0.1633} \\
    \addlinespace[4pt]

    \rowcolor{cyan!12}
    \multicolumn{10}{l}{\textit{Heart}} \\
    UNI          & 0.3452 & 0.2059 & 0.0981 & 0.3431 & 0.1876 & 0.0680 & 0.3452 & 0.2059 & 0.0876 \\
    UNI\_cls     & 0.3522 & 0.2064 & 0.0969 & 0.3480 & 0.1925 & 0.0705 & 0.3522 & 0.2053 & 0.0856 \\
    H0           & 0.3774 & 0.2185 & 0.0998 & 0.3740 & 0.2009 & 0.0691 & 0.3774 & 0.2185 & 0.0882 \\
    H0\_cls      & 0.3837 & 0.2062 & 0.0951 & 0.3805 & 0.1955 & 0.0694 & 0.3837 & 0.2062 & 0.0812 \\
    Virchow2     & \textbf{0.4499} & \textbf{0.2907} & \textbf{0.1197} & \textbf{0.4070} & 0.1469 & 0.0258 & \textbf{0.4499} & \textbf{0.2896} & 0.0848 \\
    Virchow2\_cls& 0.4015 & 0.2437 & 0.1144 & 0.4015 & \textbf{0.2318} & \textbf{0.0842} & 0.4015 & 0.2437 & \textbf{0.1007} \\
    \addlinespace[4pt]

    \rowcolor{teal!10}
    \multicolumn{10}{l}{\textit{Liver}} \\
    UNI          & 0.5084 & 0.3952 & 0.2190 & 0.5046 & 0.3853 & 0.1531 & 0.5084 & 0.3952 & 0.1658 \\
    UNI\_cls     & 0.5192 & 0.3933 & 0.2140 & 0.5116 & 0.3808 & 0.1435 & 0.5192 & 0.3933 & 0.1627 \\
    H0           & 0.5535 & 0.4169 & 0.2359 & 0.5460 & 0.4062 & 0.1684 & 0.5535 & 0.4169 & 0.1875 \\
    H0\_cls      & 0.5163 & 0.4063 & 0.2129 & 0.5163 & 0.4005 & 0.1424 & 0.5163 & 0.4063 & 0.1590 \\
    Virchow2     & \textbf{0.6840} & \textbf{0.6118} & \textbf{0.3995} & \textbf{0.6609} & \textbf{0.5165} & 0.1707 & \textbf{0.6840} & \textbf{0.5881} & \textbf{0.2234} \\
    Virchow2\_cls& 0.5540 & 0.4416 & 0.2484 & 0.5483 & 0.4305 & \textbf{0.1826} & 0.5540 & 0.4416 & 0.1919 \\
    \addlinespace[4pt]

    \rowcolor{cyan!8}
    \multicolumn{10}{l}{\textit{Bone}} \\
    UNI          & 0.1870 & 0.1114 & 0.0487 & 0.1679 & 0.0854 & 0.0242 & 0.1870 & 0.1108 & 0.0355 \\
    UNI\_cls     & 0.1764 & 0.1066 & 0.0462 & 0.1553 & 0.0798 & 0.0219 & 0.1764 & 0.1058 & 0.0330 \\
    H0           & 0.1941 & 0.1213 & 0.0538 & 0.1705 & 0.0893 & 0.0262 & 0.1941 & 0.1187 & 0.0386 \\
    H0\_cls      & 0.2036 & 0.1232 & 0.0587 & 0.1851 & 0.0937 & 0.0271 & 0.2036 & 0.1213 & 0.0419 \\
    Virchow2     & \textbf{0.2872} & \textbf{0.2189} & \textbf{0.1068} & \textbf{0.2113} & \textbf{0.1024} & \textbf{0.0290} & \textbf{0.2812} & \textbf{0.1842} & \textbf{0.0566} \\
    Virchow2\_cls& 0.2127 & 0.1204 & 0.0553 & 0.1897 & 0.0960 & 0.0282 & 0.2127 & 0.1194 & 0.0403 \\
    \addlinespace[4pt]

    \rowcolor{teal!7}
    \multicolumn{10}{l}{\textit{Brain}} \\
    UNI          & 0.4148 & 0.2012 & 0.0768 & 0.4148 & 0.1861 & 0.0558 & 0.4148 & 0.1978 & 0.0593 \\
    UNI\_cls     & 0.3969 & 0.1755 & 0.0667 & 0.3929 & 0.1626 & 0.0491 & 0.3969 & 0.1719 & 0.0515 \\
    H0           & 0.4419 & 0.2113 & 0.0787 & 0.4419 & 0.1953 & 0.0580 & 0.4419 & 0.2077 & 0.0608 \\
    H0\_cls      & 0.4284 & 0.1909 & 0.0707 & 0.4284 & 0.1772 & 0.0520 & 0.4284 & 0.1898 & 0.0585 \\
    Virchow2     & 0.4145 & \textbf{0.2949} & \textbf{0.1539} & 0.4044 & \textbf{0.2355} & \textbf{0.0731} & 0.3906 & \textbf{0.2511} & \textbf{0.0950} \\
    Virchow2\_cls& \textbf{0.4920} & 0.2338 & 0.0897 & \textbf{0.4920} & 0.2123 & 0.0636 & \textbf{0.4920} & 0.2274 & 0.0717 \\
    \addlinespace[4pt]

    \rowcolor{blue!10}
    \multicolumn{10}{l}{\textit{Ovary}} \\
    UNI          & 0.5296 & 0.3233 & 0.1559 & 0.5295 & 0.2828 & 0.0907 & 0.5296 & 0.3221 & 0.1348 \\
    UNI\_cls     & 0.5252 & 0.3201 & 0.1521 & 0.5252 & 0.2757 & 0.0879 & 0.5252 & 0.3189 & 0.1317 \\
    H0           & 0.5217 & 0.3116 & 0.1489 & 0.5217 & 0.2821 & 0.0924 & 0.5217 & 0.3098 & 0.1280 \\
    H0\_cls      & 0.5272 & 0.3115 & 0.1393 & 0.5272 & 0.2776 & 0.0820 & 0.5272 & 0.3095 & 0.1194 \\
    Virchow2     & \textbf{0.6421} & \textbf{0.5263} & \textbf{0.3106} & \textbf{0.6258} & \textbf{0.4049} & \textbf{0.1435} & \textbf{0.6406} & \textbf{0.5112} & \textbf{0.2187} \\
    Virchow2\_cls& 0.5219 & 0.3058 & 0.1394 & 0.5219 & 0.2760 & 0.0876 & 0.5219 & 0.3040 & 0.1175 \\
    \addlinespace[4pt]

    \rowcolor{gray!15}
    \multicolumn{10}{l}{\textit{Average}} \\
    UNI          & 0.4248 & 0.2662 & 0.1295 & 0.4136 & 0.2417 & 0.0864 & 0.4248 & 0.2653 & 0.1062 \\
    UNI\_cls     & 0.4142 & 0.2573 & 0.1249 & 0.4030 & 0.2331 & 0.0830 & 0.4142 & 0.2562 & 0.1027 \\
    H0           & 0.4321 & 0.2733 & 0.1336 & 0.4264 & 0.2524 & 0.0918 & 0.4321 & 0.2720 & 0.1104 \\
    H0\_cls      & 0.4253 & 0.2609 & 0.1232 & 0.4187 & 0.2415 & 0.0809 & 0.4253 & 0.2601 & 0.0996 \\
    \textbf{Virchow2} & \textbf{0.5071} & \textbf{0.3877} & \textbf{0.2165} & \textbf{0.4768} & \textbf{0.2801} & 0.0916 & \textbf{0.5018} & \textbf{0.3666} & \textbf{0.1399} \\
    Virchow2\_cls& 0.4667 & 0.2898 & 0.1397 & 0.4526 & 0.2656 & \textbf{0.0972} & 0.4667 & 0.2882 & 0.1142 \\
    \bottomrule
    \end{tabular}%
    }
\end{table}

\section{Discussion About Context Information}
Spatial transcriptomics prediction fundamentally relies on establishing precise mappings between local tissue morphology and site-specific gene expression, requiring the model to capture fine-grained features such as cellular composition, tissue architecture, and microenvironment characteristics within each patch and its spatial neighborhood. The CLS token, as a global summary of the entire image, encodes highly abstract and compressed slide-level semantics. Such overly high-level information not only struggles to provide complementary cues for local prediction, but may instead introduce noise signals that are irrelevant to the target patch, thereby interfering with the model's ability to learn accurate morphology-to-expression correspondences. This effect is especially pronounced for more powerful encoders such as Virchow2, whose patch tokens already contain sufficiently rich local semantics; appending the CLS token effectively dilutes the informative signal within the feature space, leading to marked degradation in generalization. These findings indicate that, for tasks demanding precise local reasoning such as spatial transcriptomics prediction, directly leveraging fine-grained patch-level features constitutes a more principled design choice than attempting to incorporate global information through the CLS token.

\section{Per-Cell Crop Baseline}
To isolate the effect of querying a shared token map, we trained a baseline that uses the same Virchow2 backbone, training split and gene head, but represents each cell by the CLS embedding of a $224\times224$ crop centred on the cell, computed with the frozen backbone. Table~\ref{tab:crop} compares it with CELLO under the same single-cell evaluation. CELLO is higher at every $k$ in both settings. Because CELLO also fine-tunes its backbone, the gap reflects both shared-token querying and representation adaptation.

\begin{table}[!htbp]
\centering\small
\caption{Per-cell crop baseline vs.\ CELLO (HVG PCC, single-cell evaluation).}
\label{tab:crop}
\begin{tabular}{lcccccc}
\toprule
 & \multicolumn{3}{c}{ID} & \multicolumn{3}{c}{OOD} \\
\cmidrule(lr){2-4}\cmidrule(lr){5-7}
Method & 10 & 50 & 200 & 10 & 50 & 200 \\
\midrule
Per-cell crop & 0.4981 & 0.3126 & 0.1258 & 0.3184 & 0.1691 & 0.0535 \\
CELLO         & 0.5822 & 0.3988 & 0.1629 & 0.4526 & 0.2656 & 0.0972 \\
\bottomrule
\end{tabular}
\end{table}

\section{External Validation}
We applied CELLO without adaptation to five Xenium samples from HEST-1k that are not part of the 10X-Xenium-52 (Table~\ref{tab:external}), using the same evaluation protocol.

\begin{table}[!htbp]
\centering\small
\caption{External validation on five additional HEST-1k Xenium samples (PCC at $k$ = 10 / 50 / 200).}
\label{tab:external}
\begin{tabular}{llrccc}
\toprule
Sample & Organ (subtype) & Cells & MPG & HVG & SVG \\
\midrule
TENX94  & Breast (ILC)  & 356,730 & 0.700 / 0.580 / 0.358 & 0.700 / 0.575 / 0.331 & 0.700 / 0.580 / 0.350 \\
TENX190 & Lung (LUAD)   & 278,349 & 0.580 / 0.436 / 0.234 & 0.578 / 0.421 / 0.198 & 0.580 / 0.436 / 0.220 \\
TENX199 & Breast (IDC)  & 149,995 & 0.454 / 0.314 / 0.125 & 0.454 / 0.314 / 0.125 & 0.454 / 0.314 / 0.125 \\
TENX189 & Lung (LUAD)   & 276,951 & 0.518 / 0.358 / 0.219 & 0.485 / 0.285 / 0.089 & 0.516 / 0.350 / 0.140 \\
TENX158 & Skin (SKCM)   & 111,193 & 0.604 / 0.401 / 0.199 & 0.470 / 0.196 / 0.051 & 0.604 / 0.390 / 0.119 \\
\bottomrule
\end{tabular}
\end{table}

\section{Sensitivity to Cell Localization}
\label{app:localization}
CELLO needs cell centroids at inference. To measure how localization errors affect it, we perturbed the query coordinates of a fixed random subset of 3,000 test patches without changing the ground truth, either by adding zero-mean Gaussian noise to each centroid or by randomly removing a fraction of cells. Table~\ref{tab:localization} reports the mean per-patch PCC, computed over all cells and measured genes of a patch. Performance degrades gradually and drops by less than 1\% at a 10-pixel error.

\begin{table}[!htbp]
\centering\small
\caption{Sensitivity of CELLO to cell localization errors on 3,000 test patches.}
\label{tab:localization}
\begin{tabular}{llrcc}
\toprule
Perturbation & Level & Cells & Mean PCC & Change \\
\midrule
None & -- & 43,572 & 0.1348 & -- \\
\midrule
\multirow{5}{*}{Centroid jitter (std)}
 & 5 px  & 43,572 & 0.1346 & $-0.2\%$ \\
 & 10 px & 43,572 & 0.1336 & $-0.9\%$ \\
 & 20 px & 43,572 & 0.1294 & $-4.1\%$ \\
 & 40 px & 43,572 & 0.1206 & $-10.6\%$ \\
 & 80 px & 43,572 & 0.1114 & $-17.4\%$ \\
\midrule
\multirow{3}{*}{Cell removal}
 & 10\% & 39,196 & 0.1336 & $-0.9\%$ \\
 & 20\% & 34,817 & 0.1324 & $-1.8\%$ \\
 & 40\% & 26,177 & 0.1286 & $-4.6\%$ \\
\bottomrule
\end{tabular}
\end{table}

\section{Computational Efficiency Comparison Details}
All methods were timed on one NVIDIA L40S GPU (48 GB) used exclusively, with fp16 autocast, non-overlapping $224\times224$ patches and 16 images per forward pass. Each WSI was copied to local NVMe and its page cache pre-warmed, warm-up used the same tensor shapes as the timed runs, every timed stage was bracketed by \texttt{torch.cuda.synchronize()}, and each measurement was repeated five times. For each of the 12 test slides we timed 200 randomly sampled patches and scaled the result to the whole slide by its number of patches. Inference time excludes model loading and cell segmentation; segmentation is reported separately.

Per 100 patches (mean over four slides), CELLO spends 0.253 s on reading the image region, 0.327 s on the Virchow2 forward pass (81.8\% of GPU time), 0.005 s on grid sampling, 0.053 s on cross-attention, 0.014 s on CLS fusion and 0.001 s on the gene head.

Projected whole-slide inference time is $67.5\pm48.8$ s for CELLO ($39.3\pm29.3$ s on the GPU), $73.9\pm52.1$ s for UNet3+ and $947.2\pm1068.3$ s for DeepSpot2Cell. CELLO is 14.0$\times$ faster than DeepSpot2Cell on average (median over slides 11.7$\times$, range 3.5--26.3$\times$; 21.9$\times$ in GPU time). The gap follows the number of backbone forward passes, 114,429 for CELLO versus 2,996,926 for DeepSpot2Cell. Against UNet3+, the times are comparable (median 1.12$\times$). Cell segmentation with CellViT-SAM-H takes $382.5\pm334.1$ s per slide (74--1,067 s). Including it, CELLO remains 2.5$\times$ faster than DeepSpot2Cell (median) and on par with UNet3+ (1.02$\times$).

\section{Limitations}
\label{app:limitations}
Although our benchmark covers 52 paired Xenium--WSI samples across 12 organs, several OOD organs are represented by a single slide, which limits the precision with which cross-organ generalization can be estimated. In the future work, collecting more pair data covering a wider range of tissues and organs will improve the model's generalization ability. Apart from this, in actual deployment, although we avoid masking the precise shape of cells, obtaining cell locations still relies on cell segmentation. Improving the accuracy of cell segmentation can reduce the probability of bad predictions.

CELLO requires cell locations at inference, so its predictions depend on an upstream cell detector. The OOD test set contains a single slide per unseen organ, and Kidney and Liver appear in training under other health conditions, so these results are case studies rather than estimates of organ-level generalization.


\clearpage

\end{document}